\RequirePackage{amsmath}
\documentclass{article}
\pdfoutput=1
\usepackage{iclr2021_conference,times}

\usepackage[utf8]{inputenc}
\usepackage{textcomp}
\usepackage{pgfplots}
\usepackage{pgfplotstable}
\usepackage{booktabs}
\usepackage{multirow}
\usepackage[caption=false]{subfig}
\usepackage{paralist}
\usepackage{amssymb}
\usepackage{sansmath}
\usepackage{bm}
\usepackage{mathrsfs}
\usepackage{mathdots}
\usepackage{mathtools}
\usepackage[export]{adjustbox}
\usepackage[hidelinks]{hyperref}
\usepackage[section]{placeins}

\pgfplotsset{
    compat=1.15,
    tick label style = {font=\sansmath\sffamily},
    every axis label = {font=\sansmath\sffamily},
    legend style = {font=\sansmath\sffamily},
    label style = {font=\sansmath\sffamily}
}
\graphicspath{{figures/}}

\newcommand{\SO}{\mathrm{SO}}
\newcommand{\SE}{\mathrm{SE}}
\newcommand{\E}{\mathrm{E}}

\newcommand{\pin}{\bm{p}_{\mathrm{in}}}
\newcommand{\po}{\bm{p}_{\mathrm{out}}}

\newcommand{\qi}{\bm{q}_{\mathrm{in}}}
\newcommand{\qo}{\bm{q}_{\mathrm{out}}}
\newcommand{\Dp}{\Delta\bm{p}}
\newcommand{\ci}{c_{\mathrm{in}}}
\newcommand{\co}{c_{\mathrm{out}}}
\newcommand{\cf}{c_{\mathrm{filter}}}
\newcommand{\mi}{m_{\mathrm{in}}}
\newcommand{\mip}{m'_{\mathrm{in}}}
\newcommand{\mo}{m_{\mathrm{out}}}
\newcommand{\mop}{m'_{\mathrm{out}}}
\newcommand{\mf}{m_{\mathrm{filter}}}
\newcommand{\mfp}{m'_{\mathrm{filter}}}
\newcommand{\li}{l_{\mathrm{in}}}
\newcommand{\lo}{l_{\mathrm{out}}}
\newcommand{\lf}{l_{\mathrm{filter}}}
\newcommand{\taui}{\bm{\tau}_{\mathrm{in}}}
\newcommand{\tauo}{\bm{\tau}_{\mathrm{out}}}

\begin{document}
\title{Rotation-Equivariant Deep Learning \\ for Diffusion MRI}

\author{Philip Müller $^\text{1}$ \quad Vladimir Golkov $^\text{1}$ \quad Valentina Tomassini $^\text{2}$ \quad Daniel Cremers $^\text{1}$\\
$^\text{1}$ Computer Vision Group, Technical University of Munich, Germany \\
$^\text{2}$ Department of Neurosciences, Imaging and Clinical Sciences, D'Annunzio University, \\ Chieti--Pescara, Italy \\
\texttt{\{philip.j.mueller, vladimir.golkov, cremers\}@tum.de} \\
\texttt{valentina.tomassini@unich.it}}

\maketitle 

\begin{abstract}
Convolutional networks are successful, but they have recently been outperformed by new neural networks that are equivariant under rotations and translations. These new networks work better because they do not struggle with learning each possible orientation of each image feature separately. So far, they have been proposed for 2D and 3D data. Here we generalize them to 6D diffusion MRI data, ensuring joint equivariance under 3D roto-translations in image space and the matching 3D rotations in $q$-space, as dictated by the image formation. Such equivariant deep learning is appropriate for diffusion MRI, because microstructural and macrostructural features such as neural fibers can appear at many different orientations, and because even non-rotation-equivariant deep learning has so far been the best method for many diffusion MRI tasks. We validate our equivariant method on multiple-sclerosis lesion segmentation. Our proposed neural networks yield better results and require fewer scans for training compared to non-rotation-equivariant deep learning. They also inherit all the advantages of deep learning over classical diffusion MRI methods.
Our implementation is available at \url{https://github.com/philip-mueller/equivariant-deep-dmri} and can be used off the shelf without understanding the mathematical background.

\end{abstract}

\section{Introduction and Motivation}\label{sec:motivation}

\emph{Diffusion MRI} (dMRI)~\citep{dMRI_1,dMRI_3} is an imaging technique capable of inferring properties of biological-tissue microstructure non-invasively. Each dMRI scan consists of multiple images taken with different diffusion gradients. It can be interpreted as 6D data where a coordinate (i.e.\ voxel location) in the 3D space of physical positions (called \emph{$p$-space} in our work; sometimes also called $x$-space or $(x,y,z)$-space) and a coordinate in 3D diffusion-encoding space (\emph{$q$-space}) are mapped to a signal intensity measured for that 6D coordinate.
Note that while $p$-space
is discretized to a finite Cartesian 3D grid described by the voxels of the scan, the $q$-space sampling scheme
is typically not a regular grid.

\emph{Deep learning}~\citep{dl_1,dl_2,dl_book} proved highly beneficial for dMRI~\citep{qspace_learning_1}. It is a machine-learning technique to map input features to outputs.
This mapping is done in several steps called \emph{layers}.
Each layer can separate or recombine low-level features to produce more abstract high-level features.
Layers may contain \emph{learnable parameters}.
All layers are optimized jointly using a training dataset. Thus, unlike non-learnable processing pipelines, where useful information may be discarded partially in intermediate steps, deep learning can be trained in an end-to-end manner such that all processing steps are optimized to work well together and to serve the final goal.

\emph{Convolutional neural networks (CNNs)}~\citep{dl_2,dl_book} restrict the learned mappings to be translation-equivariant
and respect locality, meaning that features are detected equally well regardless of their translation (position) and regardless of what features are present far away in the image.
These guaranteed properties of feature extraction are appropriate for many applications, including medical imaging, and improve the quality of results. This explains the success of CNNs in these applications.
If rotation-equivariance (i.e.\ detecting features well, regardless of their rotational orientation) is appropriate as well, restricting the mappings further to be translation- and rotation-equivariant improves the quality of results even more, as was shown for 2D~\citep{G_CNN,harmonicNetworks,steerableCNN,steerable_rot_2d} and 3D~\citep{G_CNN_3d,3dSteerableCNNs,TFN,cubenet,so3_equi_repr,sphericalCNN,cormorant} data.
Equivariance is formally defined in Appendix~\ref{appendix:intro_group}.

In dMRI, rotation- and translation-equivariant deep learning is appropriate because features in the image (for example properties of neural fibers) should be detected equally well regardless of their position and orientation.
A rotation and translation of an object or its parts/features in the scanner affects the image via the corresponding joint rotation in $p$- and $q$-space and the corresponding translation in $p$-space (see e.g.\ Definition~2 by~\citet{leftInvariantDiffusions}).
Therefore, deep learning for dMRI should be equivariant under joint rotations in $p$- and $q$-space, and under translations in $p$-space.
This guarantees that the position and orientation of the object and its parts in the scanner accordingly affects the position and orientation of the neural network output (if it has spatial dimensions), but does not affect other properties of the output, i.e.~does not make the network output less correct.

Our contributions are the following:
\begin{itemize}
    \item proposal of a neural network layer for 6D dMRI data respecting the equivariance properties of $p$- and $q$-space (i.e.\ adaptation of~\citet{TFN} and~\citet{3dSteerableCNNs} from 3D data to 6D dMRI data), including proofs of equivariance,
    \item efficient implementation of the proposed layer respecting the sampling properties of $p$- and $q$-space,
    \item experiments using our equivariant neural networks on a dMRI dataset for segmentation of multiple sclerosis lesions, demonstrating that our methods outperform existing methods and require less training data.
\end{itemize}

\section{Related Work}\label{sec:related_work}
\paragraph{Equivariant Deep Learning}

In the last few years there has been much effort on developing neural networks
that are equivariant under the group $\SO(3)$ of rotations or the group $\SE(3)$ of roto-translations.
Several approaches use so-called irreducible representations of $\SO(3)$ and spherical harmonics
to achieve full $\SO(3)$- or $\SE(3)$-equivariance for voxel data~\citep{3dSteerableCNNs}, for point clouds~\citep{NBody,TFN,cormorant},
or for spherical signals~\citep{conv_spherical_signals,so3_equi_repr,sphericalCNN,CG_nets}.
While these approaches use the same mathematical framework for achieving equivariance as used in the present work,
they all only consider a single 3D space, whereas the present work extends these approaches to consider two linked 3D spaces at the same time, namely physical space and $q$-space.
Methods using so-called regular representations~\citep{G_CNN_3d,cubenet} instead of irreducible representations are only suitable for discrete groups~\citep{3dSteerableCNNs},
so $\SO(3)$- or $\SE(3)$-equivariance can only be achieved approximately, by using discrete subgroups of $\SO(3)$ or $\SE(3)$.

For the 2D case, there are also prior works using irreducible representations like~\citet{harmonicNetworks} (the 2D-equivalent to~\citet{3dSteerableCNNs}),~\citet{steerableCNN}, or~\citet{steerable_rot_2d}.

In~\citet{steerable_E2} a comparison of methods for the Euclidean group $\E(2)$ and its subgroups (including $\SE(2)$ and $\SO(2)$) is presented,
and~\citet{equivariant_overview} provide a comprehensive overview of approaches for $\SE(2)$, $\SO(2)$, $\SE(3)$, and $\SO(3)$.

Besides these approaches considering rotations, there are also more general methods
for homogeneous spaces~\citep{equi_compact_cnn,equi_cnn_homog}, general manifolds~\citep{gauge_equi_CNN,bspline_CNN},
and for discrete groups~\citep{G_CNN,equi_param_sharing}.

\paragraph{Deep Learning for Diffusion MRI}
Deep learning is highly beneficial for dMRI. By avoiding suboptimal processing steps, it improves the results and allows to reduce the scan time by a factor of twelve~\citep{qspace_learning_2}. So far, it has used translation-equivariance (by using CNNs), but no rotation-equivariance.
CNNs rely on the network learning rotation-equivariance during training, e.g.\ by assuming that the training dataset already demonstrates that different orientations of features have the same meaning or using data augmentation. 
However, it was shown for 2D~\citep{G_CNN,harmonicNetworks,steerableCNN} and 3D~\citep{3dSteerableCNNs,TFN,G_CNN_3d,cormorant,cubenet,sphericalCNN} data (and here we show for 6D dMRI data) that neural networks that guarantee rotation-equivariance yield better results than neural networks that need to go through a difficult learning process to achieve imperfect equivariance.

Machine learning for dMRI has also been successfully used beyond the usual supervised setting, namely for weakly-supervised localization~\citep{Golkov-et-al-ismrm2018-global}, novelty detection~\citep{Golkov-et-al-isbi2016,Golkov-et-al-ismrm2018-novelty,vasilev2020q}, and similar scenarios~\citep{Swazinna-et-al-ismrm2019}, i.e.\ detecting diseased voxels without the need for voxel-level training labels.

Apart from the named end-to-end approaches, there are also works that only replace parts of the classical processing pipeline for dMRI by deep learning.
Some methods replace the computation of diffusion tensor images~\citep{deepDTI,deep_accelerated_DTI}
or neural fibers~\citep{deep_fiber_reconstruction}
from dMRI scans, while others use classically computed diffusion tensor images~\citep{deep_DTI_alzheimer} or neural fibers~\citep{TRAFIC}
as inputs.
These methods only replace parts of the processing pipeline for dMRI data, whereas the present work replaces the whole pipeline end-to-end, which proves more optimal for dMRI~\citep{qspace_learning_1} and is the reason for the success of deep learning in general.

There are also methods that improve image resolution in $p$-space~\citep{GAN_dMRI_super_resolution,deep_reconstruction_slice_undersampled_dMRI} 
or $q$-space~\citep{qspace_learning_1,dMRI_signal_augmnetation},
 and methods for harmonizing scans taken with different gradient strengths~\citep{deep_harmony}.

Our proposed layers can be used with all of the aforementioned tasks, with the additional advantage of offering rotation-equivariance.

\section{Methods: Roto-Translationally Equivariant Layers using Irreducible Representations}
This section describes roto-translationally equivariant linear layers based on irreducible representations (irreps). 
First, in Section~\ref{sec:equi_conv_intro}, an existing $\SE(3)$-equivariant layer \emph{for 3D data} is described. In Section~\ref{sec:theoretical_layers}, we use a similar mathematical framework to propose a novel $\SE(3)$-equivariant layer \emph{for 6D dMRI data}.
Appendix~\ref{appendix:theo_non_lin} provides some notes about the use of nonlinearities in combination with the proposed layer.
\subsection{Roto-Translationally Equivariant 
Layer for 3D Data}\label{sec:equi_conv_intro}
In~\citet{TFN} and~\citet{3dSteerableCNNs}, an $\SE(3)$-equivariant linear layer for 3D data, e.g.\ non-diffusion-weighted MRI scans, has been proposed. While in~\citet{TFN} the layer is defined for general point clouds and in~\citet{3dSteerableCNNs} it is defined for 3D images (sampled on a regular grid), the layers defined in both papers follow the same principles and theoretical derivation. In this section, we define the layer in a general way covering both variants, but follow the notation from~\citet{TFN} more closely as it is more general. Appendix~\ref{appendix:theo_background} provides some mathematical background on the building blocks of the layer, namely on groups and tensors including spherical tensors, the spherical harmonics, and the Clebsch--Gordan (CG) coefficients, and describes their properties from which the layer derives its equivariance.

This layer can be interpreted as an $\SE(3)$-equivariant analog of the well-known convolutional layer: A convolutional layer performs a so-called (multi-channel) group convolution~\citep{G_CNN} for the group of translations and thus is translation-equivariant, whereas the layer from~\citet{TFN} and~\citet{3dSteerableCNNs} is a group convolution for $\SE(3)$ and thus is
$\SE(3)$-equivariant. An approach based on the spherical harmonics basis and the
irreps of $\SO(3)$ is chosen, which achieves equivariance under $\SE(3)$ (not just under a discrete subgroup, as would be the case with regular representations, where the elements of the subgroup form a finite basis). 
Thus, a feature map, i.e.~the input (or output) of such a layer, is a multi-channel spherical-tensor field (see Appendix~\ref{appendix:tensor_field}). 
The layer uses a rotation-equivariant filter, which is also a multi-channel spherical-tensor field, and applies it  position-wise (i.e.\ for each position independently)
to the input feature map using the tensor product (which for spherical tensors includes the CG coefficients, see Appendix~\ref{appendix:CG}). 
It is built as a weighted sum, using learned weights, from predefined \emph{basis filters}.
Each of these basis filters can be decomposed multiplicatively into an angular part and a radial part.
The angular part, which depends only on directions, is given by the spherical harmonics and thus has spherical tensors as values.
The radial part, which depends only on lengths, is some set of radial basis functions and is scalar-valued. We will call the set of basis filters the \emph{filter basis}, and the sets of angular parts and radial parts \emph{angular (filter) basis} and \emph{radial (filter) basis}, respectively.

The spherical harmonics together with the CG coefficients used in the tensor product form an
angular basis of rotation-equivariant filters mapping between spherical tensors. 
Together with some radial basis, they form a complete basis for the space of rotation-equivariant linear mappings,
built by multiplying each angular basis filter with each radial basis filter~\citep{3dSteerableCNNs}.
Therefore, we can use them in basis filters to build equivariant linear mappings between multi-channel spherical-tensor fields. 
Note that the CG coefficients are required to decompose the outputs of the mapping into spherical tensors.
The complete filter basis is finite if the number of different orders in the input and output fields is finite.

A mapping from an input channel of given order $\li$ (see Appendix~\ref{appendix:tensor_field})
to an output channel of given order $\lo$ contains angular basis filters of different angular filter orders $\lf$ (where the orders $\lf$ dictate the orders of the used spherical harmonics). The orders $\lf$ can be freely chosen respecting the following condition:
\begin{align}\label{eq:CG_layer_constraint}
|\lo - \li| \le \lf \le (\lo + \li),
\end{align}
where for the angular basis to be complete, all possible orders $\lf$ have to be included.
This condition follows from the properties of the CG coefficients as defined in Eq.~\eqref{eq:CG_non_zero}.
The filter order $\lf$ can be interpreted as the frequency index, so larger $\lf$ relate to filters of higher spatial frequencies.
In addition to the filter order $\lf$ used for the angular part, the filter basis is indexed by the radial basis index $k$, which is the enumeration from $1$ to some $K$ of arbitrary, linearly independent, concentric radial basis functions.
For each path of information flow from each input channel $\ci$ to each output channel $\co$, there may exist several basis filters
with all pairwise combinations of the possible $\lf$ and $k$ values.
Note that in practice a truncated angular basis may also be used. Thus, the used filter orders $\lf$ for the paths
of information from an input channel $\ci$ to an output channel $\co$ are hyperparameters.

Following the described intuition and the definitions in~\citet{TFN}, the layer, denoted by~$\mathcal{L}$, can be defined as follows:
\begin{align}
\begin{split}
    \label{eq:TFN_layer_definition}
     \mathcal{L}_{\mo}^{(\co)}[\bm{I}](\po) \coloneqq
     & \sum_{\ci, \lf, k}W_{\ci, \co, k}^{(\lf)}
     \sum_{\substack{\mf \in \{-\lf, \dots, \lf \}, \\ \mi \in \{-\li, \dots, \li \}}}
     C_{(\lf, \mf)(\li, \mi)}^{(\lo, \mo)} \\
  & \qquad \times \sum_{\substack{\pin \in \mathbb{R}^3}}
  F_{\mf}^{(\lf, k)}(\po - \pin)I_{\mi}^{(\ci)}(\pin),
\end{split}
\end{align}
where $\bm{I}$ denotes the input feature map, $\po$ and $\pin$ are the positions in the output and input feature map, respectively, $\co$ is the index of one of the output channels, $\lo$ is shorthand for $\lo(\co)$, i.e.~the order of the output channel $\co$,
$\mo$ is the index of the components of the output spherical tensors (see Appendix~\ref{appendix:spherical_tens_harm_D}) for the given output channel number $\co$, with $-\lo \le \mo \le \lo$, the index $\ci$ goes over all input channels, $\li$ is shorthand for $l(\ci)$, i.e.~the order of the input channel given by $\ci$, $\lf$ is the filter order (frequency index) used to index the angular filter basis, $\mf$ is the index of the tensor components  (as the filter is a spherical-tensor field) of the filter of order $\lf$, with $-\lf \le \mf \le \lf$, and $k$ is the radial basis index used to index the radial basis functions,
$\bm{W}$ are learned weights, $\bm{C}$ are the CG coefficients (see Appendix~\ref{appendix:CG}), and $\bm{F}^{(\lf, k)}$ are the basis filters.
The number of output channels and their orders $\lo(\co)$ are hyperparameters.
Note that using the definition of the tensor product from Eq.~\eqref{eq:TP_spherical}, the layer could be rewritten using the tensor product explicitly:
\begin{align}
    \label{eq:TFN_layer_definition_TP}
     \mathcal{L}_{\mo}^{(\co)}[\bm{I}](\po) \coloneqq
     & \sum_{\ci, \lf, k}W_{\ci, \co, k}^{(\lf)}
     \sum_{\substack{\pin \in \mathbb{R}^3}}
     \left(\bm{F}^{(\lf, k)}(\po - \pin)\otimes \bm{I}^{(\ci)}(\pin)\right)_{\mo}^{\lo}.
\end{align}

Following~\citet{TFN} and~\citet{3dSteerableCNNs}, a basis filter $\bm{F}^{(\lf, k)}$ can be defined
as follows: 
\begin{align}\label{eq:TFN_basis_filter}
    F_{\mf}^{(\lf, k)}(\Dp)
    \coloneqq \varphi^{(k)}\left(\left\lVert\Dp\right\rVert_2\right)Y_{\mf}^{(\lf)}\left(\frac{\Dp}{\left\lVert\Dp\right\rVert_2}\right),
\end{align}
where $\Dp = \po - \pin$,  $\varphi^{(k)}: \mathbb{R}_{\ge 0} \rightarrow \mathbb{R}$ is a radial basis function,
$\bm{Y}^{(\lf)}$  are the spherical harmonics of order $\lf$, and $\left\lVert\cdot\right\rVert_2$ denoting the Euclidean norm.
Note that $\varphi^{(k)}$ may contain learnable parameters, which means that the only learnable parameters of the layer are $\bm{W}$ and the parameters in $\varphi^{(k)}$.

The layer derives its rotational equivariance from the equivariance of the spherical harmonics and the tensor product~\citep{TFN,3dSteerableCNNs}
and is also translationally equivariant as it only uses differences of positions rather than absolute positions~\citep{TFN}.

\subsection{Roto-Translationally Equivariant 
Layer for Diffusion MRI Data}\label{sec:theoretical_layers}

We will now develop a novel linear layer with special equivariance properties for use with dMRI data. To this end, we generalize the layer described in Section~\ref{sec:equi_conv_intro} and proposed in~\citet{TFN} and~\citet{3dSteerableCNNs} from acting on some 3D space to acting on the 6D space of dMRI scans.
Therefore, the feature maps, i.e.\ the inputs and outputs of the layer, are generalized from multi-channel spherical-tensor fields over a single 3D space (e.g.\ only $p$-space of MRI scans) to fields over two coupled 3D spaces, i.e.\ $p$- and $q$-space of dMRI scans, which together (by taking the direct sum of both spaces) form a 6D space.
As described in Section~\ref{sec:motivation}, the layer should be equivariant under joint rotations in $p$- and $q$-space and under translations in $p$-space.
This equivariance ensures that properties of each neural fiber such as orientation, anisotropy, orientational distribution, axon diameter, local arrangement across several centimeters, neuroplasticity, (de)myelination, or inflammation are learned and detected equally precisely, regardless of the orientation of the fiber.

When generalizing the layer, one important aspect is how the filter can be generalized to act on two 3D spaces instead of just one.
As the approach described in Section~\ref{sec:equi_conv_intro} proved quite successful for 3D data~\citep{TFN,3dSteerableCNNs}, we decided to modify it as little as possible. Therefore, as in~\citet{TFN,3dSteerableCNNs}, the filters in our proposed layer are built from a radial and an angular part, where the radial part is based on some radial basis function and the angular part is based on the spherical harmonics.
In order for the radial part to depend on both $p$- and $q$-space coordinates,
the radial basis function is applied to $p$- and $q$-space coordinates independently and the results, which do not yet depend on the image data, are combined multiplicatively.
For the angular part, we developed two approaches:
\begin{inparaenum}[i)]
\item applying the spherical harmonics to the difference of $p$- and $q$-space coordinate offsets (where \emph{offset} refers to the difference between input and output coordinates), and
\item applying the spherical harmonics to $p$- and $q$-space coordinate offsets independently and combining both results,
which do not yet depend on the image data, using the tensor product.
\end{inparaenum}
In the following sections the layer and the proposed filter bases are defined mathematically. All proofs of equivariance are postponed to Appendix~\ref{appendix:proofs}.

\subsubsection{Layer Properties and Definition}\label{sec:dmri_layer_def}
Formally, we define a feature map $\bm{I}$ over $p$- and $q$-space as a function $\bm{I} \colon \mathbb{R}^3 \oplus \mathbb{R}^3 \to \mathcal{S}^{\bm{\tau}}$, which is an extended variant
of the multi-channel spherical-tensor field described in Appendix~\ref{appendix:tensor_field}, where $\mathcal{S}^{\bm{\tau}}$ is the vector space of multi-channel spherical tensors of type $\bm{\tau}$.
An example is a usual scalar-valued 6D dMRI image (the input to the first layer).
Based on the properties of dMRI scans (Section~\ref{sec:motivation}),
a roto-translation $(g, \mathcal{T}_{\bm{t}}) \in \SE(3)$ with $g \in \SO(3)$ and $\bm{t} \in \mathbb{R}^3$ acts on such a feature map $\bm{I}$ as follows:
\begin{align}\label{eq:dMRI_feature_map_action}
   (g, \mathcal{T}_{\bm{t}})[\bm{I}](\bm{p}, \bm{q}) = \bm{D}^{\bm{\tau}}_g
   \bm{I}(\bm{\mathcal{R}}_{g^{-1}}(\bm{p}-\bm{t}), \bm{\mathcal{R}}_{g^{-1}}\bm{q}),
\end{align}
for $\bm{p} \in \mathbb{R}^3$ and $\bm{q} \in \mathbb{R}^3$.
This means that the rotation is applied to $p$- and $q$-space while the translation is only applied to the $p$-space,
which is because only $p$-space is equivariant under translations while $q$-space is not. In other words, a roto-translation of an object in the scanner rotates and translates the image in $p$-space and rotates the image in $q$-space.
We describe the proposed layer $\mathcal{L}$ as a map from the input to the output feature map:
\begin{align}
  \mathcal{L}\colon (\mathbb{R}^3 \oplus \mathbb{R}^3 \to \mathcal{S}^{\taui}) &\to
   (\mathbb{R}^3 \oplus \mathbb{R}^3 \to \mathcal{S}^{\tauo}),
\end{align}
where $\taui$ is the type (describing how many channels of which tensor orders it contains) of the input feature map  and $\tauo$ is a hyperparameter (per layer) defining the type of the output feature map.
The layer $\mathcal{L}$ should
be equivariant under roto-translations $(g, \mathcal{T}_{\bm{t}}) \in \SE(3)$ applied using Eq.~\eqref{eq:dMRI_feature_map_action}:
\begin{align}\label{eq:conv_eq_property}
   \left(\mathcal{L} \circ (g, \mathcal{T}_{\bm{t}})\right)[\bm{I}]\left(\bm{p}, \bm{q}\right) =
   \left((g, \mathcal{T}_{\bm{t}}) \circ \mathcal{L}\right)[\bm{I}]\left(\bm{p}, \bm{q}\right).
\end{align}
In other words, applying a roto-translation $(g, \mathcal{T}_{\bm{t}})$ to the input feature map should have the same result as applying it to the output feature map.

\begin{figure}[t]
  \centering
  \includegraphics[width=1.0\textwidth]{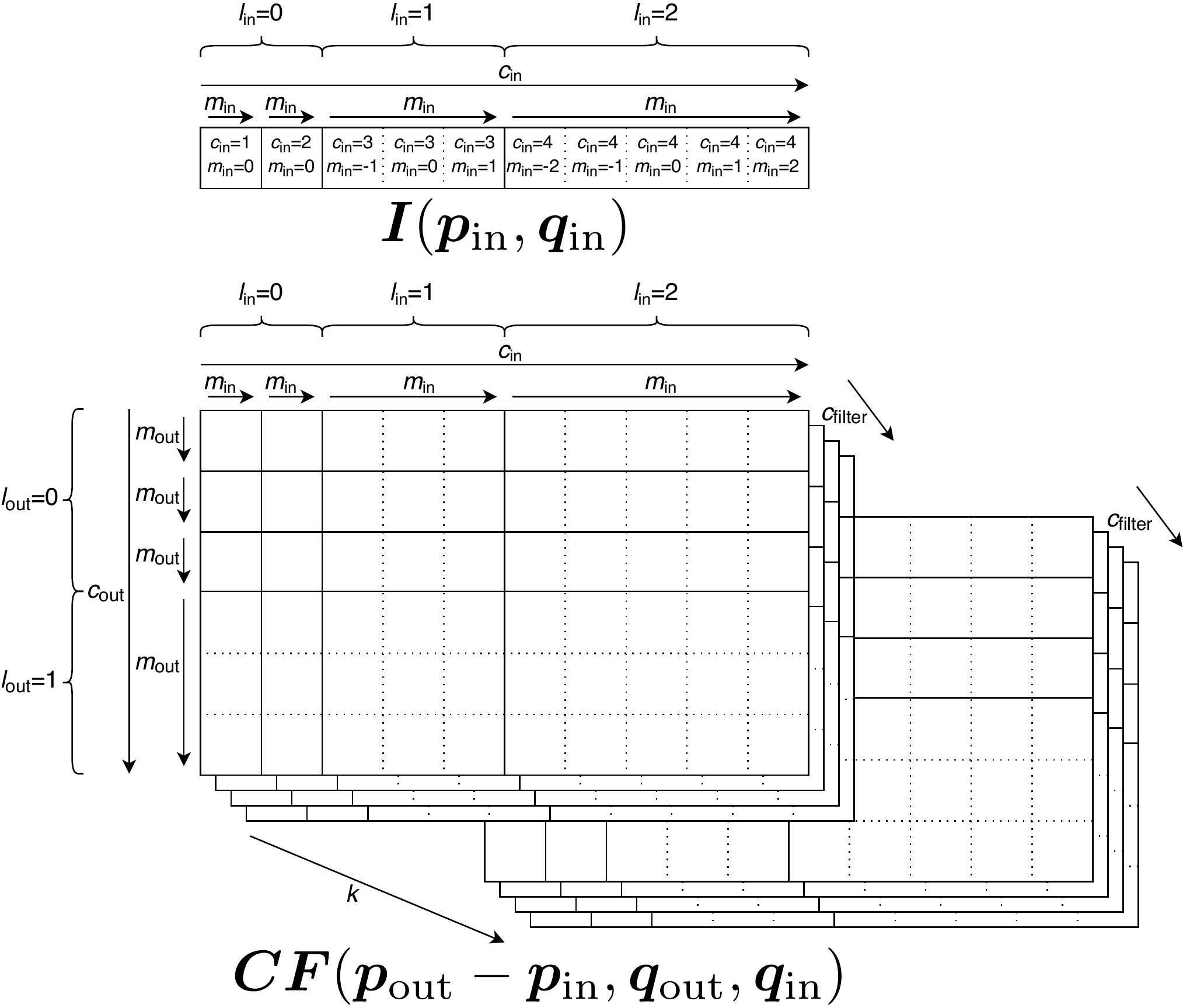}
  \caption{Example structure of the elements used in Eq.~\eqref{eq:conv_layer_definition} when using two scalar ($\li=0$), one vector ($\li=1$), and one $\li=2$ input channels and three scalar ($\lo=0$) and one vector ($\lo=1$) output channels. The input features $\bm{I}(\pin, \qi)$ at point $(\pin, \qi)$ are a multi-channel spherical tensor consisting of multiple concatenated spherical tensors, the channels, indexed by $\ci$, where the components of each channel are indexed by $\mi$. A spherical tensor of order $l$ has $2l+1$ components (see Appendix~\ref{appendix:spherical_tens_harm_D}), so $\bm{I}(\pin, \qi)$ in this example has $\sum_{\ci} 2l(\ci)+1 = 2 \cdot 1 + 1 \cdot 3 + 1 \cdot 5 = 10$ entries. The notation $\bm{C}\bm{F}(\po - \pin, \qo, \qi)$ is shorthand for the set $\left\{\sum_{\mf \in \{-\lf, \dots, \lf \}}
     C_{(\lf, \mf)(\cdot, \cdot)}^{(\cdot, \cdot)} F_{\mf}^{(\cf, k)}(\po - \pin, \qo, \qi)\right\}_{\cf, k}$ over the indices $\cf, k$, which is the filter $\bm{F}$ at position $(\po - \pin, \qo, \qi)$ combined with the CG coefficients $\bm{C}$. Each of the elements in this set is a matrix with the same number of columns as $\bm{I}(\pin, \qi)$ and the same number of rows as the resulting output spherical tensor $\bm{\mathcal{L}}[\bm{I}](\po, \qo)$. Thus, $\bm{C}$ is applied linearly to the filter $\bm{F}$ and transforms it into a set of linear mappings from the input to the output multi-channel spherical tensors, where there is a linear mapping for each angular filter channel $\cf$ and radial basis index $k$.
     The linear mapping for each pair $\cf,k$ contributes to the output by filtering the input with specific angular resolutions and radii.
  }
  \label{fig:filter_formula_elements}
\end{figure}
\begin{figure}[t]
  \centering
  \includegraphics[width=1.0\textwidth]{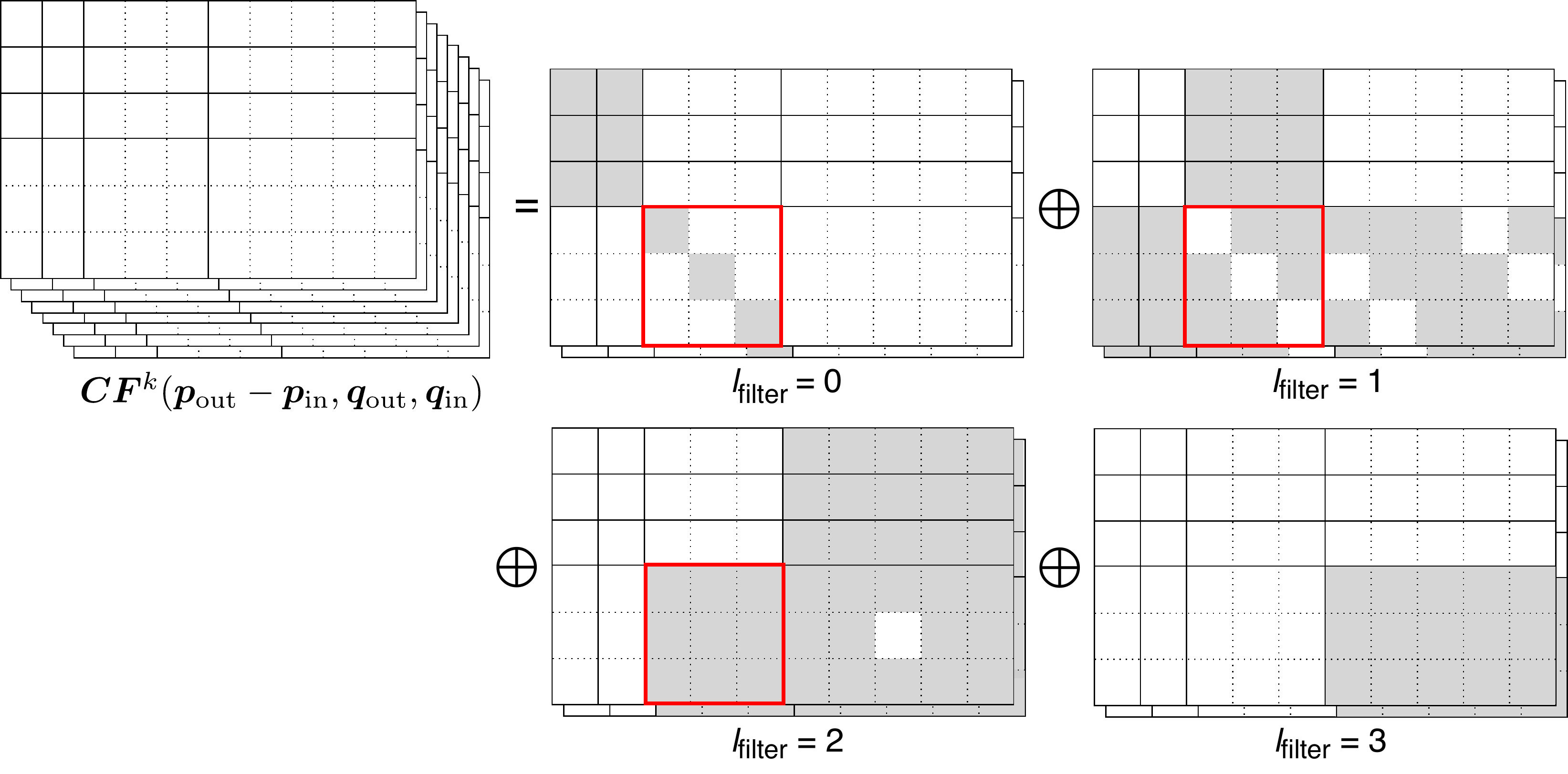}
  \caption{Structure of the angular filter basis combined with $\bm{C}$. For a given radial basis index $k$, the filter $\bm{F}$ at position $(\po - \pin, \qo, \qi)$ combined with $\bm{C}$ (see Figure~\ref{fig:filter_formula_elements}) can be decomposed into groups of angular filter channels $\cf$ of same order $\lf$, where $\oplus$ denotes the direct sum (concatenation). Note that the number of filter channels, i.e.~possible values of $\cf$, for each order $\lf$ depends on the used angular basis. Thus, the shown stacked filter channels are meant to show that there might be multiple channels and should not be interpreted as the exact number of channels.
  If the angular basis filters are the spherical harmonics, then there is only one $\cf$ for each $\lf$. For each $\lf$, only specific elements can be non-zero (highlighted in grey), which follows from the properties of the CG coefficients.
  For $\lf=0$, the sections for $\lo=0, \li=0$ represent the scalar-scalar product and the section for $\lo=1, \li=1$ represents the vector-scalar product (each output component is based on the same input component) of input and filter. In $\lf=1$, the $\lo=0, \li=1$ sections represent the dot product (the output is based on all components of the input), the $\lo=1, \li=0$ sections represent the scalar-vector product (the input influences all output components), and the $\lo=1, \li=1$ section represents the cross product (each input component influences all other output components but not the one at the same index). Figure~\ref{fig:SH_C} explains how the exact values in the red squares are obtained from the values $(\po - \pin, \qo, \qi)$ if the used angular basis are the spherical harmonics applied to one 3D space, e.g.~$p$-space offsets as in Eq.~\eqref{eq:angular_basis_p}, $q$-space offsets as in Eq.~\eqref{eq:angular_basis_q}, or the space of differences between $p$- and $q$-space offsets in Eq.~\eqref{eq:angular_basis_pq}.}
  \label{fig:angular_C}
\end{figure}
\begin{figure}[t]
  \centering
  \includegraphics[width=1.0\textwidth]{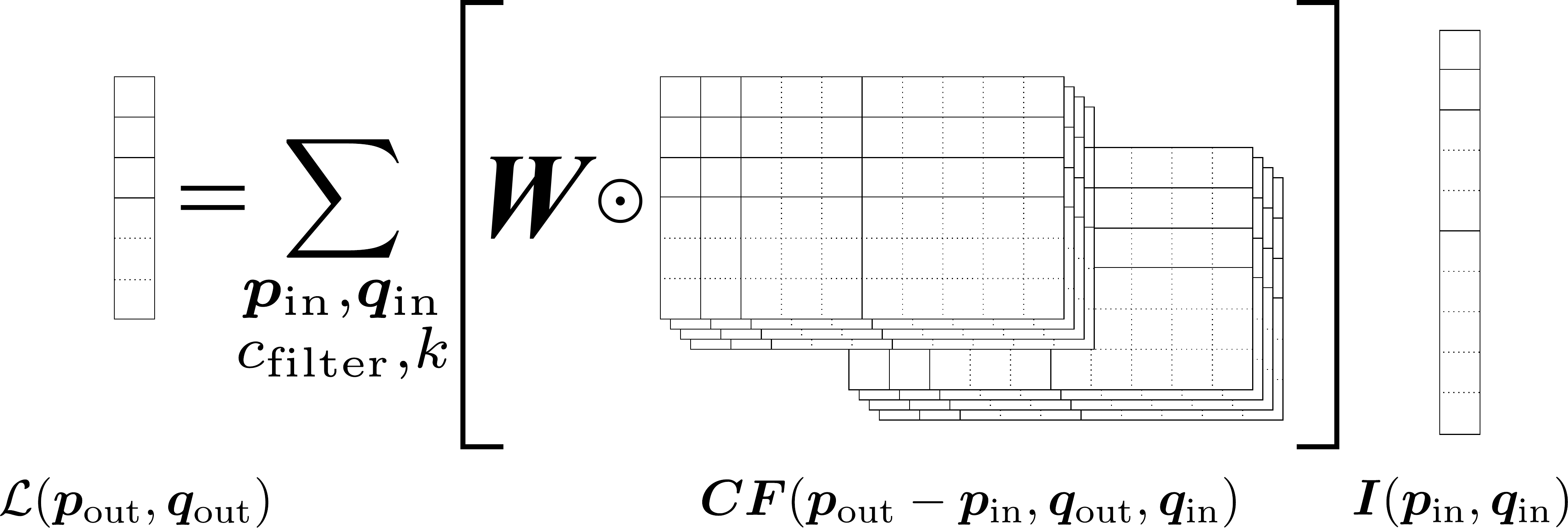}
  \caption{Interpretation of Eq.~\eqref{eq:conv_layer_definition}. 
  At point $\po, \qo$ and for each possible $\pin, \qi$, first
  $\bm{C}\bm{F}(\po - \pin, \qo, \qi)$ is built (see Figure~\ref{fig:filter_formula_elements}). 
  Then it is multiplied element-wise, denoted by $\odot$, with the trainable weights array $\bm{W}$. The result is matrix-multiplied (for each $\cf, k$ independently) with $\bm{I}(\pin, \qi)$. Finally, we sum over all $\pin, \qi, \cf, k$.}
  \label{fig:filter_formula}
\end{figure}

In order to define the proposed layer, for which Eq.~\eqref{eq:conv_eq_property} holds,
Eq.~\eqref{eq:TFN_layer_definition} is generalized by adding dependence on $q$-space coordinates and replacing the angular filter order $\lf$ by the more general angular filter channel $\cf$:
\begin{align}\label{eq:conv_layer_definition}
\begin{split}
     \mathcal{L}_{\mo}^{(\co)}[\bm{I}](\po, \qo) \coloneqq
     &\sum_{\ci, \cf, k}W_{\ci, \co, k}^{(\cf)}
     \sum_{\substack{\mf \in \{-\lf, \dots, \lf \}, \\ \mi \in \{-\li, \dots, \li \}}}
     C_{(\lf, \mf)(\li, \mi)}^{(\lo, \mo)}\\
  & \qquad \times \sum_{\substack{\pin \in \mathbb{R}^3, \\ \qi \in \mathbb{R}^3}}
  F_{\mf}^{(\cf, k)}(\po - \pin, \qo, \qi)I_{\mi}^{(\ci)}(\pin, \qi),
\end{split}
\end{align}
where $\qo$ and $\qi$
are $q$-space coordinates in the output and input feature map, respectively, $\cf$ is the angular filter channel index used to index the angular filter basis, $\lf$ is shorthand for $l(\cf)$, i.e.\ the filter order of the angular filter channel given by $\cf$, $\bm{F}^{(\cf, k)}$ are the basis filters, and the other variables and symbols are
as in Eq.~\eqref{eq:TFN_layer_definition}.
We introduced the angular filter channel $\cf$ in order to allow generalization of the basis filters. While the filter basis defined in Eq.~\eqref{eq:TFN_basis_filter} contains only a single basis filter for given $k$ and $\lf$,
we allow it to contain multiple of them so that there may be multiple angular filter channels $\cf$ for a given $\lf$, which is important for supporting filters built using the tensor product as we propose it.
Like in Eq.~\eqref{eq:TFN_layer_definition}, the decision which $\lf$ values to use for each path of information from a $\ci$ to a $\co$ is a hyperparameter that can be freely chosen respecting Eq.~\eqref{eq:CG_layer_constraint}. The number of angular filter channels $\cf$ for each $\lf$ then depends on the chosen angular basis and its hyperparameters.
Note that following~\citet{TFN} the basis filters only depend on position differences as this is sufficient for translational equivariance in $p$-space, but we allow them to treat $\qo$ and $\qi$ independently as no translational equivariance is required in $q$-space.
Figure~\ref{fig:filter_formula_elements} shows the intuitive structure of important elements of Eq.~\eqref{eq:conv_layer_definition} for the example hyperparameters $\taui=(2,1,1), \tauo=(3,1,0), \cf\in\{1,2,3,4\}, k\in\{1,2\}$. Figure~\ref{fig:angular_C}
shows some details
of the angular basis combined with $\bm{C}$.
Figure~\ref{fig:filter_formula} gives an interpretation of Eq.~\eqref{eq:conv_layer_definition} using the elements shown in Figure~\ref{fig:filter_formula_elements}.
Eq.~\eqref{eq:conv_layer_definition} uses the definition of the tensor product from Eq.~\eqref{eq:TP_spherical} and, analogously to Eq.~\eqref{eq:TFN_layer_definition_TP}, the layer could instead be defined using the tensor product explicitly:
\begin{align}
\begin{split}
    \label{eq:conv_layer_definition_TP}
     &\mathcal{L}_{\mo}^{(\co)}[\bm{I}](\po, \qo) \\
     & \qquad \coloneqq
     \sum_{\ci, \cf, k}W_{\ci, \co, k}^{(\cf)}
     \sum_{\substack{\pin \in \mathbb{R}^3, \\ \qi \in \mathbb{R}^3}}
     \Big(\bm{F}^{(\cf, k)}(\po - \pin, \qo, \qi) \otimes \bm{I}^{(\ci)}(\pin, \qi)\Big)_{\mo}^{\lo}.
\end{split}
\end{align}

We require that all basis filters $\bm{F}^{(\cf, k)}$ are equivariant under joint rotations:
\begin{align}\label{eq:kernel_eq_property}
  \bm{F}^{(\cf, k)}(\bm{\mathcal{R}}_g\Dp, \bm{\mathcal{R}}_g\qo, \bm{\mathcal{R}}_g\qi)
= \bm{D}^{(\lf)}_g\bm{F}^{(\cf, k)}(\Dp, \qo, \qi),
\end{align}
which is an extension of the $\SO(3)$-equivariance required for Eq.~\eqref{eq:TFN_basis_filter}, and implies the required equivariance of the layer, as proved in Appendix~\ref{appendix:layer_equi}.

\subsubsection{General Filter Basis}\label{sec:conv_filter_kernel}
Each basis filter $\bm{F}^{(\cf, k)}\colon \left(\mathbb{R}^3\right)^3 \to \mathcal{S}^{\lf}$ maps the tuple $(\Dp, \qo, \qi) \in \left(\mathbb{R}^3\right)^3$ (the position difference $\Dp = \po - \pin$
and the two $q$-space coordinates $\qo$ and $\qi$)
to a spherical tensor of order $\lf$. This filter tensor will later be applied (using the tensor product) to the input feature map.
As in Eq.~\eqref{eq:TFN_basis_filter}, the basis filter consists of a radial basis and an angular basis, which we generalize to functions $R^{(k)}\colon \left(\mathbb{R}^3\right)^3 \to \mathbb{R}$
and $\bm{A}^{(\cf)}\colon \left(\mathbb{R}^3\right)^3 \to \mathcal{S}^{\lf}$, respectively.
Thus, $\bm{F}^{(\cf, k)}$ is defined as
\begin{align}
  \bm{F}^{(\cf, k)}(\Dp, \qo, \qi)
  \coloneqq R^{(k)}(\Dp, \qo, \qi)
  \bm{A}^{(\cf)}(\Dp, \qo, \qi).
  \label{eq:def_general_filter_kernel}
\end{align}
We require $R$ to be invariant under rotations:
\begin{align}
  R^{(k)}(\bm{\mathcal{R}}_g\Dp, \bm{\mathcal{R}}_g\qo, \bm{\mathcal{R}}_g\qi) = R^{(k)}(\Dp, \qo, \qi) \qquad \forall g \in \SO(3),
  \label{eq:R_invariance}
\end{align}
and $\bm{A}$ to be equivariant under rotations:
\begin{align}
  \bm{A}^{(\cf)}(\bm{\mathcal{R}}_g\Dp, \bm{\mathcal{R}}_g\qo, \bm{\mathcal{R}}_g\qi) =
  \bm{D}^{(\lf)}_g \bm{A}^{(\cf)}(\Dp, \qo, \qi) \qquad \forall g \in \SO(3),
  \label{eq:A_equivariance}
\end{align}
as this is sufficient for $\bm{F}$~\eqref{eq:def_general_filter_kernel} to satisfy Eq.~\eqref{eq:kernel_eq_property} as is proven in Appendix~\ref{appendix:filter_basis_equi}.
Various options for $R$ and $\bm{A}$ are described in Section~\ref{sec:conv_radial_basis} and Section~\ref{sec:conv_angular_basis}, respectively, 
and in Section~\ref{sec:filter_kernel_types} we propose basis filters built using these options.

\subsubsection{Radial Filter Basis}\label{sec:conv_radial_basis}
We use the following simple forms for the radial basis:
\begin{itemize}
    \item a radial basis using only $\Dp$, as in Eq.~\eqref{eq:TFN_basis_filter}:
        \begin{align}\label{eq:R_p_diff}R_{p\text{-diff}}^{(k)}(\Dp, \qo, \qi) \coloneqq \varphi^{(k)}\left(\left\lVert\Dp\right\rVert_2\right),\end{align}
    \item a radial basis using only $q$-space coordinates of the input feature map, a proposed adaption of Eq.~\eqref{eq:R_p_diff}:
        \begin{align}\label{eq:R_q_in}R_{q\text{-in}}^{(k)}(\Dp, \qo, \qi) \coloneqq \varphi^{(k)}\left(\left\lVert\qi\right\rVert_2\right),\end{align}
    \item and a radial basis using only $q$-space coordinates of the output feature map, a proposed adaption of Eq.~\eqref{eq:R_p_diff}:
        \begin{align}\label{eq:R_q_out}R_{q\text{-out}}^{(k)}(\Dp, \qo, \qi) \coloneqq \varphi^{(k)}\left(\left\lVert\qo\right\rVert_2\right),\end{align}
\end{itemize}
where $\varphi^{(k)} \colon \mathbb{R}_{\ge 0} \to \mathbb{R}$ is a set of radial basis functions, e.g.\ Gaussian radial basis functions~\citep{3dSteerableCNNs}, cosine radial basis functions~\citep{e3nn}, or a fully connected network applied to a set of basis functions~\citep{TFN,e3nn}. For details on the radial basis functions, see Appendix~\ref{appendix:radial_basis_fn}. 

We propose to combine multiple radial bases multiplicatively:
\begin{align}\label{eq:R_prod}
  R_{\text{prod}}^{(k)}(\Dp, \qo, \qi) = R_{\text{prod}}^{(k_1, k_2)}(\Dp, \qo, \qi) \coloneqq
  R^{(k_1)}_1(\Dp, \qo, \qi) R^{(k_2)}_2(\Dp, \qo, \qi),
\end{align}
where $R^{(k_1)}_1$ and $R^{(k_2)}_2$ are the two radial bases being combined and each value assumed by $k$ represents one of the possible combinations of the radial basis indices $k_1$ and $k_2$ of the two combined radial bases.
This means that the radial basis size $K$ is the product of the sizes $K_1, K_2$ of the two combined radial bases: $K=K_1 K_2$.

\subsubsection{Angular Filter Basis}\label{sec:conv_angular_basis}
\begin{figure}[t]
  \centering
  \includegraphics[width=.9\textwidth]{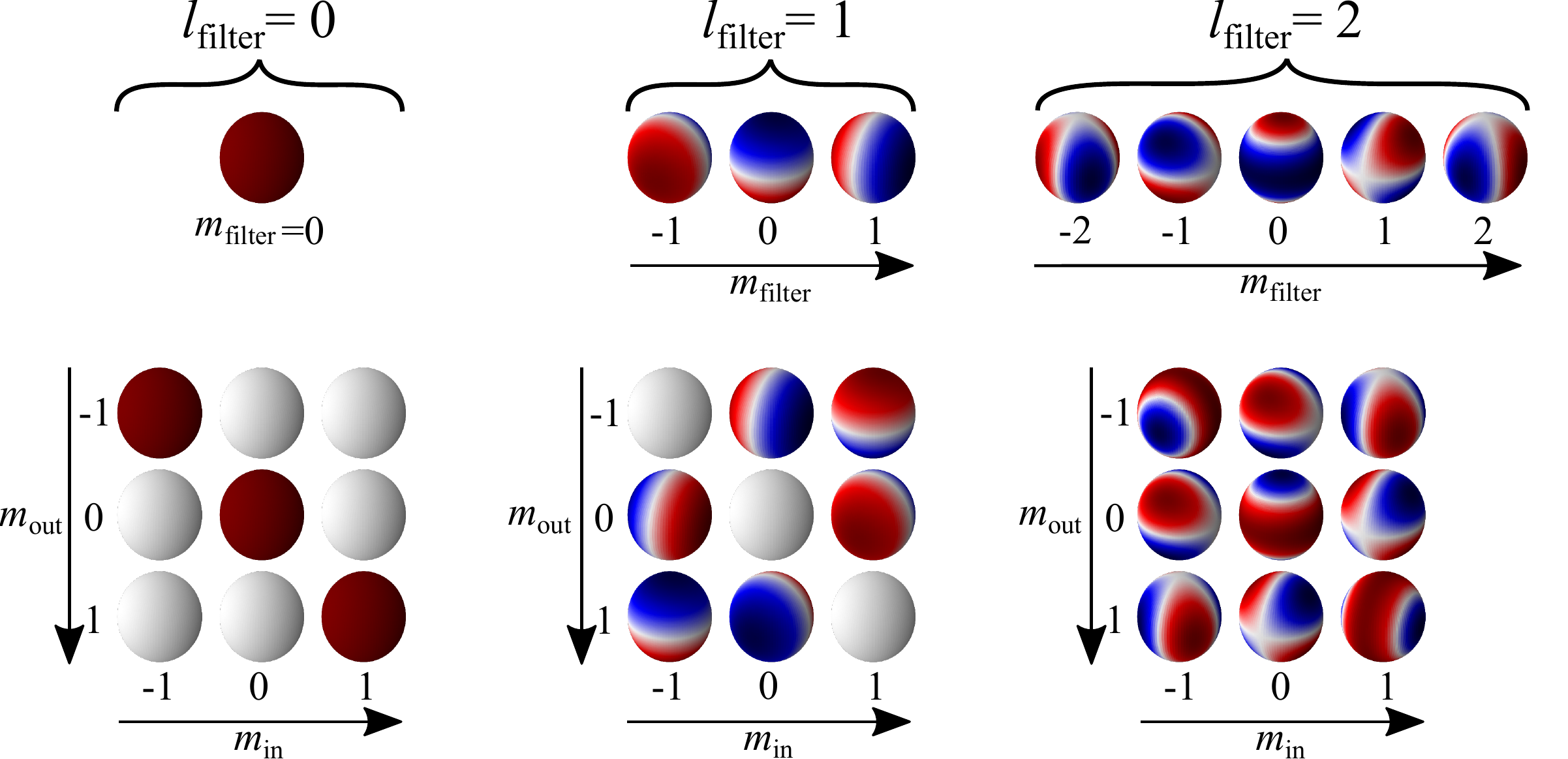}
  \caption{Visualization of the spherical harmonics and their combination with the CG coefficients. As the domain of the spherical harmonics is the sphere, they can be represented using a colored sphere for each $\lf, \mf$, where the point on the sphere represents the direction of the vector given to the spherical harmonic, i.e.~the direction of $\po - \pin$ in $\bm{A}_{p\text{-diff}}$, of $\qo - \qi$ in $\bm{A}_{q\text{-diff}}$, and of $(\po - \qo) - (\pin - \qi)$ in $\bm{A}_{pq\text{-diff}}$, respectively. In the upper part of the figure, the spherical harmonics $Y_{\mf}^{\lf}$ of orders $\lf=0, 1, 2$ are shown. When used as angular filters for vector input and output ($\li=\lo=1$), they can be combined with the CG coefficients $\bm{C}_{0, 1}^1$, $\bm{C}_{1, 1}^1$, and $\bm{C}_{2, 1}^1$, respectively.
  The results of this combination are shown in the lower part of the figure. By applying these angular basis functions to the aforementioned direction vector, we obtain the three $3\times 3$ matrices in the ($\li=\lo=1$) section of the filter, shown as red squares in Figure~\ref{fig:angular_C}.}
  \label{fig:SH_C}
\end{figure}
We use the following angular bases, all based on the (real) spherical harmonics $\bm{Y}$ (see Appendix~\ref{appendix:spherical_tens_harm_D}):
\begin{itemize}
\item an angular basis using only $\Dp$, as in Eq.~\eqref{eq:TFN_basis_filter}:
    \begin{align}\label{eq:angular_basis_p}
      \bm{A}^{(\cf)}_{p\text{-diff}}(\Dp, \qo, \qi) \coloneqq
      \bm{Y}^{(\lf)}\left(\frac{\Dp}{\left\lVert\Dp\right\rVert_2}\right) = \bm{Y}^{(\lf)}_{\mf}\left(\frac{\po - \pin}{\left\lVert\po - \pin\right\rVert_2}\right),
    \end{align}
    
\item an angular basis using only the $q$-difference, i.e.~the offset between input and output $q$-space coordinates, a proposed adaption of Eq.~\eqref{eq:angular_basis_p}:
    \begin{align}\label{eq:angular_basis_q}
      \bm{A}^{(\cf)}_{q\text{-diff}}(\Dp, \qo, \qi) \coloneqq
      \bm{Y}^{(\lf)}\left(\frac{\qo - \qi}{\left\lVert\qo - \qi\right\rVert_2}\right),
    \end{align}

\item and an angular basis using the $pq$-difference,  i.e.~the difference between the input/output offsets of $p$- and $q$-space coordinates, which is the same as the input/output offsets of the differences between $p$- and $q$-space coordinates,
a proposed adaption of Eq.~\eqref{eq:angular_basis_p}:
    \begin{align}\label{eq:angular_basis_pq}
    \begin{split}
      \bm{A}^{(\cf)}_{pq\text{-diff}}(\Dp, \qo, \qi) \coloneqq {}
      &\bm{Y}^{(\lf)}\left(\frac{\Dp - (\qo - \qi)}{\left\lVert\Dp - (\qo - \qi)\right\rVert_2}\right) \\
      = {}
      &\bm{Y}^{(\lf)}\left(\frac{(\po - \qo) - (\pin - \qi)}{\left\lVert(\po - \qo) - (\pin - \qi)\right\rVert_2}\right),
    \end{split}
    \end{align}
\end{itemize}
where $\lf$ is shorthand for $l(\cf)$, i.e.~the filter order of the angular filter channel given by $\cf$.
The intuition of $\bm{A}_{pq\text{-diff}}$ is that the spherical harmonics expect a 3D unit vector as input and we want it to depend on $p$- and $q$-space coordinates, so both coordinates need to be combined to another 3D unit vector in an equivariant way. We choose the difference operation as it is a very simple linear operation.
Figure~\ref{fig:SH_C} visualizes the spherical harmonics and their combination with the CG coefficients. As the introduced angular bases $\bm{A}_{p\text{-diff}}$, $\bm{A}_{q\text{-diff}}$, and $\bm{A}_{pq\text{-diff}}$ are all based on the spherical harmonics, this figure also provides some intuition about these bases.

\begin{table}[t]
    \centering
    \caption{Visualization of the angular basis $\bm{A}_\text{TP}$, Eq.~\eqref{eq:angular_combined_def}, for the filter order $\lf = 0, 1$, and the orders $l_1 = 0, 1$ and $l_2 = 0,1$ of the angular bases $\bm{A_1}$ and $\bm{A_2}$ being combined.
    As the bases $\bm{A_1}$ and $\bm{A_2}$ are not defined specifically, i.e.\ they may be any function with specific properties, $\bm{A}_\text{TP}$ cannot be visualized in general. Therefore, we visualize a specific variant of $\bm{A}_\text{TP}$ where $\bm{A_1}$ and $\bm{A_2}$ are the spherical harmonics applied to $p$- and $q$-space coordinate offsets, respectively, i.e.\ $l_1=l_p$, $l_2=l_q$, $\left(\bm{A}_1\right)^{(l_1)}=\bm{Y}^{(l_p)}\left(\frac{\Dp}{\left\lVert\Dp\right\rVert_2}\right)$, $\left(\bm{A}_2\right)^{(l_2)} = \bm{Y}^{(l_q)}\left(\frac{\qo - \qi}{\left\lVert\qo - \qi\right\rVert_2}\right)$.
    With this definition, $\bm{A}_\text{TP}$ corresponds to the angular part of $\bm{F}_\text{TP}$ as defined in Eq.~\eqref{eq:F_TP}.
    The domain of $\bm{A_1}$, $\bm{A_2}$, and $\bm{A}_\text{TP}$ consists of 6D directions, i.e.\ the directions of the two 3D vectors $\Dp$ and $(\qo - \qi)$.
    We project it to 3D for visualization purposes by plotting the azimuthal angle from the first 3D space ($\Dp$) and the polar angle from the second 3D space $(\qo - \qi)$ and fixing the other angles to constant values. Thus, even if plotted spheres are equal, the non-projected values may not be.
    In the rows with $\lf=1$ and $l_1=0$ or $l_2=0$ it can be seen that the resulting $\bm{A}_\text{TP}$ only depends on $\bm{A_2}$ (i.e.\ the content of the blue rectangles is identical) or $\bm{A_1}$ (i.e.\ the content of the orange rectangles is identical), respectively. This is because $\left(\bm{A}_1\right)^{(0)}$ and $\left(\bm{A}_2\right)^{(0)}$ are scalars and thus independent of directions.
    }\label{tab:angular_basis_TP}
    \newcommand{\includekernelgraphic}[1]{\includegraphics[scale=.5,trim=10 10 10 10,clip]{#1}}
\newcommand{\includekernelgraphiccolor}[2]{\includegraphics[scale=.5,trim=10 10 10 10,clip,cfbox=#2 1pt 1pt]{#1}}
\begin{tabular}{>{\centering\arraybackslash}m{.8cm}>{\centering\arraybackslash}m{.5cm}>{\centering\arraybackslash}m{.5cm}>{\centering\arraybackslash}m{3cm}>{\centering\arraybackslash}m{3cm}>{\centering\arraybackslash}m{3cm}}
    \toprule \\
     \multirow{2}{*}{$\lf$} & $l_1$ & $l_2$ & $\left(\bm{A}_1\right)^{(l_1)}$ & $\left(\bm{A}_2\right)^{(l_2)}$ & $\left(\bm{A}\right)_\text{TP}^{(\lf, l_1, l_2)}$ \\
      & $l_p$ & $l_q$ & $ \bm{Y}^{(l_p)}\left(\frac{\Dp}{\left\lVert\Dp\right\rVert_2}\right)$  & $\bm{Y}^{(l_q)}\left(\frac{\qo - \qi}{\left\lVert\qo - \qi\right\rVert_2}\right)$  & $ \bm{F}_\text{TP}^{(\lf, l_p, l_q)}$  \\
     &&& \tiny $m_1=-1 \qquad 0 \qquad +1$ & \tiny $m_2=-1 \qquad 0 \qquad +1$ & \tiny $\mf=-1 \qquad 0 \qquad +1 \qquad$ \\
     \midrule \\
     0 & 0 & 0 & \includekernelgraphic{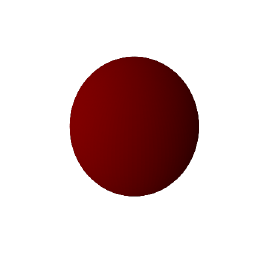} & \includekernelgraphic{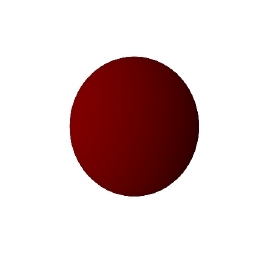} & \includekernelgraphic{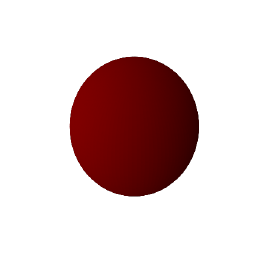} \\
     0 & 1 & 1 & \includekernelgraphic{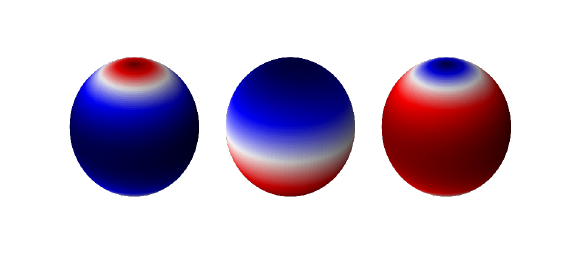} & \includekernelgraphic{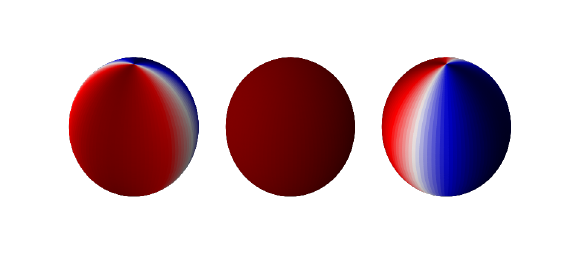} & \includekernelgraphic{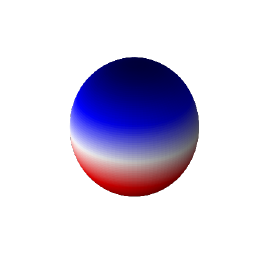} \\
     1 & 0 & 1 & \includekernelgraphic{figures/single_space_kernel_1_0.pdf} & \includekernelgraphiccolor{figures/single_space_kernel_2_1.pdf}{blue} & \includekernelgraphiccolor{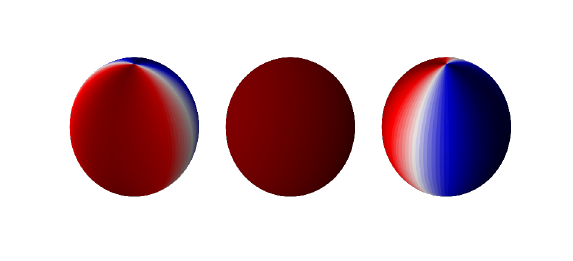}{blue} \\
     1 & 1 & 0 & \includekernelgraphiccolor{figures/single_space_kernel_1_1.pdf}{orange} & \includekernelgraphic{figures/single_space_kernel_2_0.pdf} & \includekernelgraphiccolor{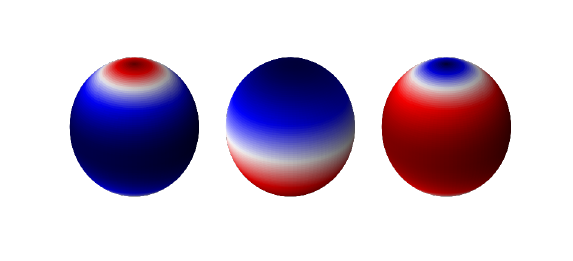}{orange} \\
     1 & 1 & 1 & \includekernelgraphic{figures/single_space_kernel_1_1.pdf} & \includekernelgraphic{figures/single_space_kernel_2_1.pdf} & \includekernelgraphic{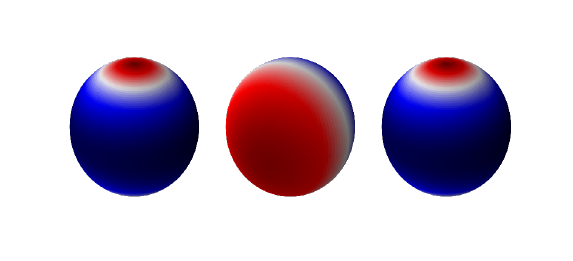} \\
     \bottomrule
\end{tabular}
\end{table}
Instead of combining the $p$- and $q$-space coordinates before applying the spherical harmonics, we also propose to combine two angular bases $\left(\bm{A}_1\right)^{(c_1)}$ and $\left(\bm{A}_2\right)^{(c_2)}$,
which may depend on $p$- and $q$-space coordinates, respectively, using the tensor product (as defined in Eq.~\eqref{eq:TP_spherical}):
\begin{align}\label{eq:angular_combined_def}
\begin{split}
    \left(A_\text{TP}\right)^{(\cf)}_{\mf}(\Dp, \qo, \qi)
    = &\left(A_\text{TP}\right)^{(\lf, c_1, c_2)}_{\mf}(\Dp, \qo, \qi) \\
  \coloneqq &\sum_{\substack{m_1 \in \{-l_1, \dots, l_1\}, \\ m_2 \in \{-l_2, \dots, l_2\}}}
  C_{(l_1, m_1)(l_2, m_2)}^{(\lf, \mf)} \\
  & \qquad \times \left(A_1\right)^{(c_1)}_{m_1}(\Dp, \qo, \qi)
  \left(A_2\right)^{(c_2)}_{m_2}(\Dp, \qo, \qi),
\end{split}
\end{align}
where each value assumed by $\cf$ represents one  combination of filter order $\lf$ and angular basis channels $c_1$, $c_2$ of the angular bases $\bm{A}_1,\bm{A}_2$ that are being combined, $l_1, l_2$ denote the orders these channels, and $\mf$ is the
index of the components of the spherical tensors produced by the filter for given $\cf$, with $-\lf \le \mf \le \lf$.
When building filter channels of a given filter order $\lf$, the orders $l_1, l_2$ can be freely chosen as long as they satisfy the following constraint, which follows from the properties of the Clebsch--Gordan coefficients as defined in Eq.~\eqref{eq:CG_non_zero}:
\begin{align}\label{eq:TP_filter_CG_constraint}
  |l_1 - l_2| \le \lf \le (l_1 + l_2).
\end{align}
Table~\ref{tab:angular_basis_TP} visualizes the angular basis $\bm{A}_\text{TP}$ and provides some intuition.

\subsubsection{Proposed Filter Bases}\label{sec:filter_kernel_types}
Using the angular and radial parts discussed in Section~\ref{sec:conv_radial_basis} and Section~\ref{sec:conv_angular_basis}, various variants of filter bases
can be built.
As proven in Appendix~\ref{appendix:filter_basis_equi}, the equivariance of these filter bases follows from the
invariance of their radial parts, proven in Appendix~\ref{appendix:proof_radial_basis_inv}, and the equivariance of their angular parts, proven in Appendix~\ref{appendix:proof_angular_basis_equi}.
In this work, the following filter bases are proposed:
\paragraph{$p$-Space Filter} 
  The following filter basis depends only on $p$-space coordinates and is built using the $p$-difference radial basis~\eqref{eq:R_p_diff} and the $p$-difference angular basis~\eqref{eq:angular_basis_p}:
      \begin{align}\label{eq:F_p}
      \begin{split}
        F_{p\text{-space}, \mf}^{(\cf, k)}(\Dp, \qo, \qi)
        ={} & F_{p\text{-space}, \mf}^{(\lf, k)}(\Dp, \qo, \qi) \\
        \coloneqq {} &\varphi^{(k)}(\left\lVert\Dp\right\rVert_2)Y_{\mf}^{(\lf)}\left(\frac{\Dp}{\left\lVert\Dp\right\rVert_2}\right).
      \end{split}
      \end{align}

\paragraph{$q$-Space Filter}
    The following filter basis depends only on $q$-space coordinates and is built using the multiplicative combination~\eqref{eq:R_prod} of the $q$-in~\eqref{eq:R_q_in} and $q$-out~\eqref{eq:R_q_out} radial bases, and using the $q$-difference angular basis~\eqref{eq:angular_basis_q}:
      \begin{align}\label{eq:F_q}
      \begin{split}
        F_{q\text{-space}, \mf}^{(\cf, k)}(\Dp, \qo, \qi)
        ={} & F_{q\text{-space}, \mf}^{(\lf, k_1, k_2)}(\Dp, \qo, \qi) \\
        \coloneqq{} &\varphi^{(k_1)}(\left\lVert\qo\right\rVert_2)\varphi^{(k_2)}(\left\lVert\qi\right\rVert_2)
        Y_{\mf}^{(\lf)}\left(\frac{\qo - \qi}{\left\lVert\qo - \qi\right\rVert_2}\right).
      \end{split}
      \end{align}
      
 \paragraph{$pq$-difference Filter}
    The following filter basis depends on $p$- and $q$-space coordinates and is built using the multiplicative combination~\eqref{eq:R_prod} of the $p$-difference~\eqref{eq:R_p_diff}, $q$-in~\eqref{eq:R_q_in}, and $q$-out~\eqref{eq:R_q_out} radial bases, and using the $pq$-difference angular basis~\eqref{eq:angular_basis_pq}:
    \begin{align}\label{eq:F_pq_diff}
    \begin{split}
        F_{pq\text{-diff}, \mf}^{(\cf, k)}(\Dp, \qo, \qi)
        ={} & F_{pq\text{-diff}, \mf}^{(\lf, k_1, k_2, k_3)}(\Dp, \qo, \qi) \\
        \coloneqq{} &\varphi^{(k_1)}(\left\lVert\Dp\right\rVert_2)
        \varphi^{(k_2)}(\left\lVert\qo\right\rVert_2)\varphi^{(k_3)}(\left\lVert\qi\right\rVert_2) \\
        &\qquad \times Y_{\mf}^{(\lf)}\left(\frac{\Dp - (\qo - \qi)}{\left\lVert\Dp - (\qo - \qi)\right\rVert_2}\right).
    \end{split}
    \end{align}     
      
\paragraph{Tensor Product
of $p$- and $q$-Space Filters}
    The following filter basis depends on $p$- and $q$-space coordinates and is built using the multiplicative combination~\eqref{eq:R_prod} of the $p$-difference~\eqref{eq:R_p_diff}, $q$-in~\eqref{eq:R_q_in}, and $q$-out~\eqref{eq:R_q_out} radial bases, and using the tensor product combination~\eqref{eq:angular_combined_def} of the $p$-difference~\eqref{eq:angular_basis_p} and $q$-difference~\eqref{eq:angular_basis_q} angular bases:
    \begin{align}\label{eq:F_TP}
    \begin{split}
        F_{\text{TP}, \mf}^{(\cf, k)}(\Dp, \qo, \qi)
        ={} & F_{\text{TP}, \mf}^{(\lf, l_p, l_q, k_1, k_2, k_3)}(\Dp, \qo, \qi) \\
        \coloneqq {} &\varphi^{(k_1)}(\left\lVert\Dp\right\rVert_2)
        \varphi^{(k_2)}(\left\lVert\qo\right\rVert_2)\varphi^{(k_3)}(\left\lVert\qi\right\rVert_2) \\
        &\qquad \times \sum_{\substack{m_p \in \{-l_p, \dots, l_p\}, \\ m_q \in \{-l_q, \dots, l_q\}}}
        C_{(l_p, m_p)(l_q, m_q)}^{(\lf, \mf)} \\
        &\qquad \times Y_{m_p}^{(l_p)}\left(\frac{\Dp}{\left\lVert\Dp\right\rVert_2}\right)
        Y_{m_q}^{(l_q)}\left(\frac{\qo - \qi}{\left\lVert\qo - \qi\right\rVert_2}\right),
  \end{split}
  \end{align} where $l_p,l_q$ (inserted for $l_1,l_2$ in Eq.~\eqref{eq:angular_combined_def})
  are the orders of the $p$- and $q$-space filters, i.e.\ the orders of the spherical harmonics applied to $p$- and $q$-space, respectively, and given the filter order $\lf$, may be selected freely respecting Eq.~\eqref{eq:TP_filter_CG_constraint} (with $l_p, l_q$ inserted for $l_1, l_2$).
  
Thus, given an input of type $\taui$ (the numbers of input channels of each order $\li$),
the hyperparameter selection process is as follows: 1.~select output type $\tauo$ (the numbers of output channels of each order $\lo$),
2.~for each pair $(\ci,\co)$, choose what $\lf$ values to use (respecting Eq.~\eqref{eq:CG_layer_constraint}), 3.~in the case of the TP filter, choose what $(l_p, l_q)$ tuples to use for each $\lf$ (respecting Eq.~\eqref{eq:TP_filter_CG_constraint} with $l_p, l_q$ inserted for $l_1, l_2$).

Each tuple $(\lf, l_p, l_q)$ represents a specific operation (as per Eq.~\ref{eq:TP_spherical}) used to combine the $p$- and $q$-space filters, e.g.\ the cross product in the case of $(1, 1, 1)$ or the dot product for $(0, 1, 1)$.
As for a given $\lf$ there may be multiple possible options for $l_p$ and $l_q$, the angular basis may contain several basis filters for each $\lf$. 

There may be many different strategies to choose the set of $(\lf, l_p, l_q)$. We focus on analyzing the effect of the angular basis size and thus choose two strategies, one with a small set of $(\lf, l_p, l_q)$, which thus has very few angular basis filters, and the other with a larger set and thus more basis filters.

In the first strategy we focus on operations on vectors, as vectors are the lowest-order tensors that can represent directions
and as such enable us to consider directions with minimum possible effort.
Thus, we use the cross product and scalar products, represented by tuples $(\lf, l_p, l_q)$ with values $(1, 1, 1)$, $(1, 0, 1)$, and $(1, 1, 0)$.
To also support basis filters of order $\lf=0$ and $\lf=2$, respectively, which are the neighboring filter orders of $\lf=1$, the tuples $(0, 0, 0)$ and $(2, 2, 2)$ are included.
Note that this hyperparameter choice is free and we might have also chosen to not include them or to use different $l_p, l_q$.
The filter basis using $F_{\text{TP}}$~\eqref{eq:F_TP} with the described strategy will be called $F_{\text{TP-vec}}$.

In the other strategy used to create a larger angular basis, certain rules are defined regarding which $(\lf, l_p, l_q)$ to select.
Besides constraint~\eqref{eq:TP_filter_CG_constraint},
we choose $(\lf, l_p, l_q)$ such that $l_p$ and $l_q$ do not deviate from the given $\lf$ more than by an integer hyperparameter $d$, meaning that $|\lf - l_p| \le d$
and $|\lf - l_q| \le d$. 
While this rule is arbitrary to create filter bases much larger than $F_{\text{TP-vec}}$ while still restricting the number of basis filters to a finite number, which is not restricted by constraint~\eqref{eq:TP_filter_CG_constraint}, the intuition behind this rule is to create filters from  $p$- and $q$-space filters of orders $l_p, l_q$ close to the resulting filter order $\lf$.
We call the filter basis based on $F_{\text{TP}}$~\eqref{eq:F_TP} with this strategy applied $F_{\text{TP}\pm d}$, e.g.\ $F_{\text{TP}\pm 1}$ for $d=1$.

\subsubsection{Comparison and Combinations of the Proposed Filter Bases}\label{sec:filter_basis_options}

The filter bases $F_{p\text{-space}}$~\eqref{eq:F_p} and $F_{q\text{-space}}$~\eqref{eq:F_q} each only depend on coordinates of one space, $p$- and $q$-space, respectively, thus they are invariant
under rotations and translations in the other space.
While some invariances might be wanted for the whole network, it may lead to drawbacks if
early layers are invariant. Later layers extract higher-level features and if early layers are invariant
then some information required for the higher-level features might be lost.

The $F_{pq\text{-diff}}$ basis~\eqref{eq:F_pq_diff} depends on coordinates from both spaces.
But it discards parts of the structural information of the input (not the image values), as its angular part only depends on the difference of coordinate offsets of the two spaces, which means that the structural information from coordinates in both 3D spaces is represented by only a single 3D vector.

To solve this problem, the $F_{pq\text{-diff}}$ basis may be combined with the $F_{p\text{-space}}$ or the $F_{q\text{-space}}$ basis by filtering the feature maps with both filters independently
and then combining the results by summing.
(Summing is like using more filter channels.)
These filter bases will be called $F_{pq\text{-diff}+p}$ and $F_{pq\text{-diff}+q}$.
As $F_{pq\text{-diff}}$ and $F_{p\text{-space}}$ depend on different 3D vectors which together contain all relevant structural information, $F_{pq\text{-diff}+p}$ does not discard parts of the structural information of the input. The same is true for the combination of $F_{pq\text{-diff}}$ and $F_{q\text{-space}}$ to $F_{pq\text{-diff}+q}$.

The filter basis $F_{\text{TP}}$~\eqref{eq:F_TP}, and its variants $F_{\text{TP-vec}}$ and $F_{\text{TP}\pm d}$, use structural information from both $p$- and $q$-space by combining angular basis filters of these two spaces using multiple operations, defined by the tensor product, and thus may access more aspects of the structural dependencies between both spaces than filters using $F_{pq\text{-diff}}$, which only uses as single operation, the difference. 
As the angular part of $F_{\text{TP}}$~\eqref{eq:F_TP}, in contrast to all other bases like $F_{pq\text{-diff}+p}$, may contain several basis filters for each $\lf$,
it has the largest number of parameters (as the $\bm{W}$ is larger if the basis contains more basis filters), most computational effort, and highest memory requirements for the same number and orders of input and output channels. 

\subsubsection{Implementation of the Layer}
Details on the implementation of the layer are given in Appendix~\ref{appendix:implementation}. Our code is available at \url{https://github.com/philip-mueller/equivariant-deep-dmri#}.
In order to use the layer with dMRI scans, i.e.\ with finite Cartesian $p$-space and finite sampling schemes in $q$-space, Eq.~\eqref{eq:conv_layer_definition} needs to be discretized as explained in Appendix~\ref{appendix:impl_discretization}. We follow~\citet{3dSteerableCNNs}, where the precomputation of parts of the filter is proposed for a computationally efficient and hardware-optimized implementation on voxel grids, and implement Eq.~\eqref{eq:conv_layer_definition} using 3D convolutional layers as explained in Appendix~\ref{appendix:impl_3d_conv}.

\section{Experimental Setup}
\label{sec:experiments}
The effectivity of the proposed layer was studied by doing segmentation (i.e.~voxel-wise classification) of multiple sclerosis (MS) lesions
using a dataset~\citep{cardiff_dataset,cardiff_dataset_2} containing dMRI brain scans with ground-truth annotations of MS lesions.
\subsection{Dataset and Preprocessing}
The dMRI scans are sampled at 46 $q$-space points: six times at $\bm{q}=\bm{0}$, and at 40 uniformly distributed diffusion directions ($b=1200\mathrm{s/mm}^2$, SE-EPI, voxel size $1.8\mathrm{mm}\times 1.8\mathrm{mm}\times 2.4\mathrm{mm}$, matrix $128\times 128$, $57$ slices,
TE=94.5ms, TR=16s, motion/distortion-corrected with elastix~\citep{elastix} with upsampling to $256\times 256\times 172$).
For well-behaved neural network training, so-called \emph{feature scaling} was performed by dividing each channel by the corresponding channel mean taken over all scans. To prevent overfitting on intensity values, each scan was additionally divided by its mean intensity. 
The ground truth of each sample describes for each voxel whether it contains any MS lesion or not, so it has the same resolution as the scan
but does not contain different $q$-space points.

Due to the long training times, no cross-validation was performed.
Instead, the dataset (94 MS patients) was split only once into training and validation set at an 80/20 ratio.

As the directed $q$-vectors slightly differ between the scans (due to motion correction), the mean of each of these $q$-vectors over all training samples
was used as the input $q$-sampling scheme of the network (exact $q$-vectors could be considered, but would require precomputing more filters or recomputing larger parts of the filters in each iteration).

\subsection{Network Architecture}
\label{sec:exp_architecture}
A simple architecture is chosen that first combines all $\bm{q}=\bm{0}$ channels by computing their mean,
then applies multiple of the proposed equivariant layers on $p$- and $q$-space
(they will be referred to as $pq$-layers), followed by a global reduction operation (called $q$-reduction) that ``collapses'' $q$-space and only leaves $p$-space,
and then applies multiple of the proposed equivariant layers on $p$-space (called $p$-layers).

\paragraph{$pq$-layers}
In this work the following filter bases for the $pq$-layers are investigated (see Section~\ref{sec:filter_basis_options} for details):
$F_{pq\text{-diff}+p}$, $F_{pq\text{-diff}+q}$, $F_{\text{TP-vec}}$, and $F_{\text{TP}\pm 1}$.

\paragraph{$q$-reduction}
There are several options how the $q$-space can be ``collapsed'' but in this work the comparison is focused
on the following two configurations:
\begin{itemize}
    \item \texttt{late}: In all $pq$-layers the same $q$-sampling scheme as in the input data is used.
      The $q$-space is then ``collapsed'' using a \texttt{q-length weighted average} layer,
      which applies radial basis functions on the lengths of the $q$-vectors in the sampling scheme
      and weights the results using learned weights.
    \item \texttt{gradual}: Each $pq$-layer uses a different output $q$-sampling scheme $\mathcal{Q}_\mathrm{out}$ that consists of less
      $q$-vectors than in the layer before.
      The final $q$-reduction is done using the same filtering as in the $pq$-layers
      but with $\mathcal{Q}_\mathrm{out} = \{(0,0,0)\}$.
\end{itemize}

\paragraph{$p$-layers}
As in the $p$-layers no $q$-space is present anymore, only the $F_{p\text{-space}}$~\eqref{eq:F_p} filter basis is used.
Note that this is equivalent to the layer defined in Eq.~\eqref{eq:TFN_layer_definition}.

\paragraph{Further Configuration}
Various channel configurations used in this work are defined in Appendix~\ref{appendix:equi_architectures}.
Swish~\citep{swish_act} is used as acitvation function for scalar channels and the gated nonlinearity~\citep{3dSteerableCNNs} is used for $l>0$ channels
(see Appendix~\ref{appendix:theo_non_lin}).
In $q$-space we use Gaussian radial basis functions~\eqref{eq:gauss_radial_fn} and 
in $p$-space we either use cosine radial basis functions~\eqref{eq:cosine_radial_fn} or Gaussian radial basis functions and either apply three fully connected (FC) layers, having 50 neurons each, to them (cosine+fc/Gaussian+fc) or use them alone (cosine/Gaussian).
In the kernels, all possible orders $\lf$ (see~Eq.~\eqref{eq:CG_layer_constraint}) of angular basis filters are used and the kernel size in $p$-space is set to $5$.

\subsection{Training}
Binary cross-entropy is used as loss and the sigmoid function is used as activation function in the final layer.
Using the brain masks of each sample, all voxels outside of the brain are ignored
when computing the loss and the quality metrics.
To counteract class imbalance, positive and negative voxels were weighted in the loss according to the ratio between positive and negative voxels in the whole training set.

Due to the large feature maps, much GPU memory was required during training.
To reduce the required memory to a minimum, only a batch size of one was used and each sample was cropped
to the bounding box defined by its brain mask.
Additionally, checkpointing was applied, where only some feature maps at defined checkpoints are stored
in each forward pass and the other ones are recomputed during the backward pass.
This further reduces the GPU memory consumption but increases the training time.

\subsection{Experiments}
The following research questions should be answered in this work:
\begin{inparaenum}[i)]
    \item How do the equivariant models perform compared to similar non-rotation-equivariant reference models?
    \item How do equivariant and non-equivariant models behave when the training dataset is reduced?
    \item What are the effects of filter types, $q$-reduction, and channel setups?
    \item How do the training times and memory requirements of the equivariant models compare to the non-equivariant models?
\end{inparaenum}

To answer these questions, models with different $q$-reduction strategies (\texttt{late}, \texttt{gradual}), $pq$-filter bases (\texttt{pq-diff+p}, \texttt{pq-diff+q}, \texttt{TP-vec}, \texttt{TP$\pm 1$}), layer and channel confiurations,
and radial basis functions (cosine or Gaussian with or without fully connected layers) were trained.
Furthermore, various non-rotation-equivariant reference architectures with normal 3D convolutional
layers and ReLU activation functions were trained using the same training setup and kernel sizes to be as comparable as possible.
The channel settings of these architectures are shown
in Appendix~\ref{appendix:reference_architectures}. We will refer to these models as the \emph{non-equivariant} models.

In order to answer question ii), the best equivariant and non-equivariant models were
additionally trained on subsets of different sizes of the training set but validated against the full validation set.

\section{Results and Discussion}\label{sec:results_discussion}
\paragraph{Comparison with Non-Rotation-Equivariant Reference Models}

\newcommand*\lineB[2][c]{\begin{tabular}[#1]{@{}c@{}}#2\end{tabular}}

\begin{table}[t]
    \centering
    \caption{
        Comparison of network architectures. Numbers of layers of equivariant networks are a sum of $pq$-layers, $q$-reduction layer (always one), and $p$-layers. Abbreviations are explained in Section~\ref{sec:exp_architecture} and exact hyperparameter values for channels and layers are given in Appendix~\ref{appendix:architectures}.
        All equivariant models outperform all 
        non-rotation-equivariant models (\texttt{non-eq}).
        We experimented with different numbers of channels but found that often small changes in the number of channels did not affect the quality much. For each layer configuration we only show the best models we found by hyperparameter tuning of channels and learning rates. We thus assume that they are very near the optimum for the given training setup. This is also true for the \texttt{non-eq} models (where we additionally show multiple of the best channels configurations for each number of layers).
        }\label{tab:results}
    \scriptsize
    \setlength{\tabcolsep}{3pt}
\begin{tabular}{@{}lllllllll@{}}
    \toprule
     ID & $q$- & Filter & Layers & $p$ Radial & \#params & AUC  & Avg- & Dice \\ 
    & Reduction & basis & & basis & & & Prec & score 
    \\\midrule

    l\_TP1\_1+2 & \texttt{late} & \texttt{TP$\pm 1$} & 1+1+2 & cosine+fc & 461344 & 0.9787 & 0.6088 & 0.5783 \\
    
    l\_TP1\_1+3 & \texttt{late} & \texttt{TP$\pm 1$} & 1+1+3 & cosine+fc & 585114 & \underline{\textbf{0.9820}} & 0.6198 & 0.5889 \\

     l\_TP1\_1+4 & \texttt{late} & \texttt{TP$\pm 1$} & 1+1+4 & cosine+fc & 662398 & 0.9817 & \underline{\textbf{0.6299}} & \underline{\textbf{0.5918}} \\
   l\_TP1\_1+4($l2$) & \texttt{late} & \texttt{TP$\pm 1$} & 1+1+4($l2$) & cosine+fc & 671887 & 0.9794 & 0.6033 & 0.5840 \\
   l\_TP1\_1+4($l3$) & \texttt{late} & \texttt{TP$\pm 1$} & 1+1+4($l2/l3$) &  cosine+fc & 674795 & 0.9757 & 0.6171 & 0.5890 \\
     
     l\_TP1\_1($l2$)+4($l2$) & \texttt{late} & \texttt{TP$\pm 1$} & 1($l2$)+1+4($l2$) & cosine+fc & 655748 & 0.9792 & 0.6044 & 0.5781 \\
     l\_TP1\_1($l3$)+4($l3$) & \texttt{late} & \texttt{TP$\pm 1$} & 1($l2/l3$)+1+4($l2/l3$) & cosine+fc & 802396 & 0.9753 & 0.5898 & 0.5666 \\

    l\_TPvec\_1+4 & \texttt{late} & \texttt{TP-vec} & 1+1+4 & cosine+fc & 367998 & 0.9775 & 0.6072 & 0.5777 \\
    l\_pq-diff-p\_1+4 & \texttt{late} & {\tiny \texttt{pq-diff+p}} & 1+1+4 & cosine+fc & 271509 & 0.9781 & 0.5980 & 0.5722 \\
    l\_pq-diff-q\_1+4 & \texttt{late} & {\tiny \texttt{pq-diff+q}} & 1+1+4 & cosine+fc & 269449 & 0.9798 & 0.6150 & 0.5832 \\
    
    l\_TP1\_1+4\_Gfc & \texttt{late} & \texttt{TP$\pm 1$} & 1+1+4 & Gaussian+fc & 662398 & 0.9794 & 0.5949 & 0.5657 \\
   l\_TP1\_1+4\_c & \texttt{late} & \texttt{TP$\pm 1$} & 1+1+4 & cosine & 39184 & 0.9739 & 0.5816 & 0.5656 \\
   l\_TP1\_1+4\_G & \texttt{late} & \texttt{TP$\pm 1$} & 1+1+4 & Gaussian & 39184 & 0.9702 & 0.5369 & 0.5327 \\
   
   l\_TP1\_1+5 & \texttt{late} & \texttt{TP$\pm 1$} & 1+1+5 & cosine+fc & 510594 & 0.9807 & 0.6184 & 0.5882 \\

    g\_TP1\_0+3 & \texttt{gradual} & \texttt{TP$\pm 1$} & 0+1+3 & Gaussian & 54951  & 0.9637 & 0.4661 & 0.4794 \\
    
    g\_TP1\_1+2 & \texttt{gradual} & \texttt{TP$\pm 1$} & 1+1+2 & cosine+fc & 329640  & 0.9761 & 0.5701 & 0.5552 \\

    g\_TP1\_1+3 & \texttt{gradual} & \texttt{TP$\pm 1$} & 1+1+3 & cosine+fc & 355968  & 0.9666 & 0.5301 & 0.5203 \\
    
    g\_TP1\_2+1 & \texttt{gradual} & \texttt{TP$\pm 1$} & 2+1+1 & cosine+fc & 133579  & 0.9742 & 0.5933 & 0.5713 \\

    \midrule
    n\_3\_few & - & \texttt{non-eq} & 3 {\tiny(few channels)} & - &    31009 & 0.9094 & 0.1433 & 0.2310 \\
    n\_3\_many & - & \texttt{non-eq} & 3 {\tiny(many channels)} & - &    64391 & \textbf{0.9556} & 0.4429 & 0.4745 \\
    n\_4\_few & - & \texttt{non-eq} & 4 {\tiny(few channels)} & - &    97899 & 0.9536 & \textbf{0.4540} & \textbf{0.4846} \\
    n\_4\_many & - & \texttt{non-eq} & 4 {\tiny(many channels)} & - &   216921 & 0.9532 & 0.4532 & 0.4765 \\
    n\_5\_few & - & \texttt{non-eq} & 5 {\tiny(few channels)} & - &    49779 & 0.9442 & 0.3964 & 0.4409 \\
    n\_5\_many & - & \texttt{non-eq} & 5 {\tiny(many channels)} & - &   622820 & 0.9371 & 0.3319 & 0.3907 \\
    n\_6\_few & - & \texttt{non-eq} & 6 {\tiny(few channels)} & - &    52909 & 0.9470 & 0.3699 & 0.4159 \\
    n\_6\_many & - & \texttt{non-eq} & 6 {\tiny(many channels)} & - &   116936 & 0.9238 & 0.2691 & 0.3265 \\
    n\_6\_fm\_small & - & \texttt{non-eq} & 6 {\tiny(matched fm small)} & - & 12720208 & 0.8554 & 0.0554 & 0.0517 \\
    n\_6\_fm\_large & - & \texttt{non-eq} & 6 {\tiny(matched fm large)} & - & 19590724 & 0.8671 & 0.0709 & 0.0837 \\
    \bottomrule
\end{tabular}
\end{table}
Table~\ref{tab:results} shows the results of equivariant models with different hyperparameters and the trained non-rotation-equivariant models.
It can be seen that all shown equivariant models outperform the reference models in the \emph{receiver operating characteristic (ROC)},
measured by the \emph{area under the curve (AUC)} of the ROC,
the \emph{precision-recall curve}, measured by the \emph{average precision (AvgPrec)} score, and the Dice score.
Our best equivariant model outperforms the best non-rotation-equivariant model
by $2.9 \%$ 
in AUC, by $38.7 \%$ 
in AvgPrec, and by $22.1 \%$ in Dice score.

\begin{figure}[t]
  \centering
  \includegraphics[width=.9\textwidth]{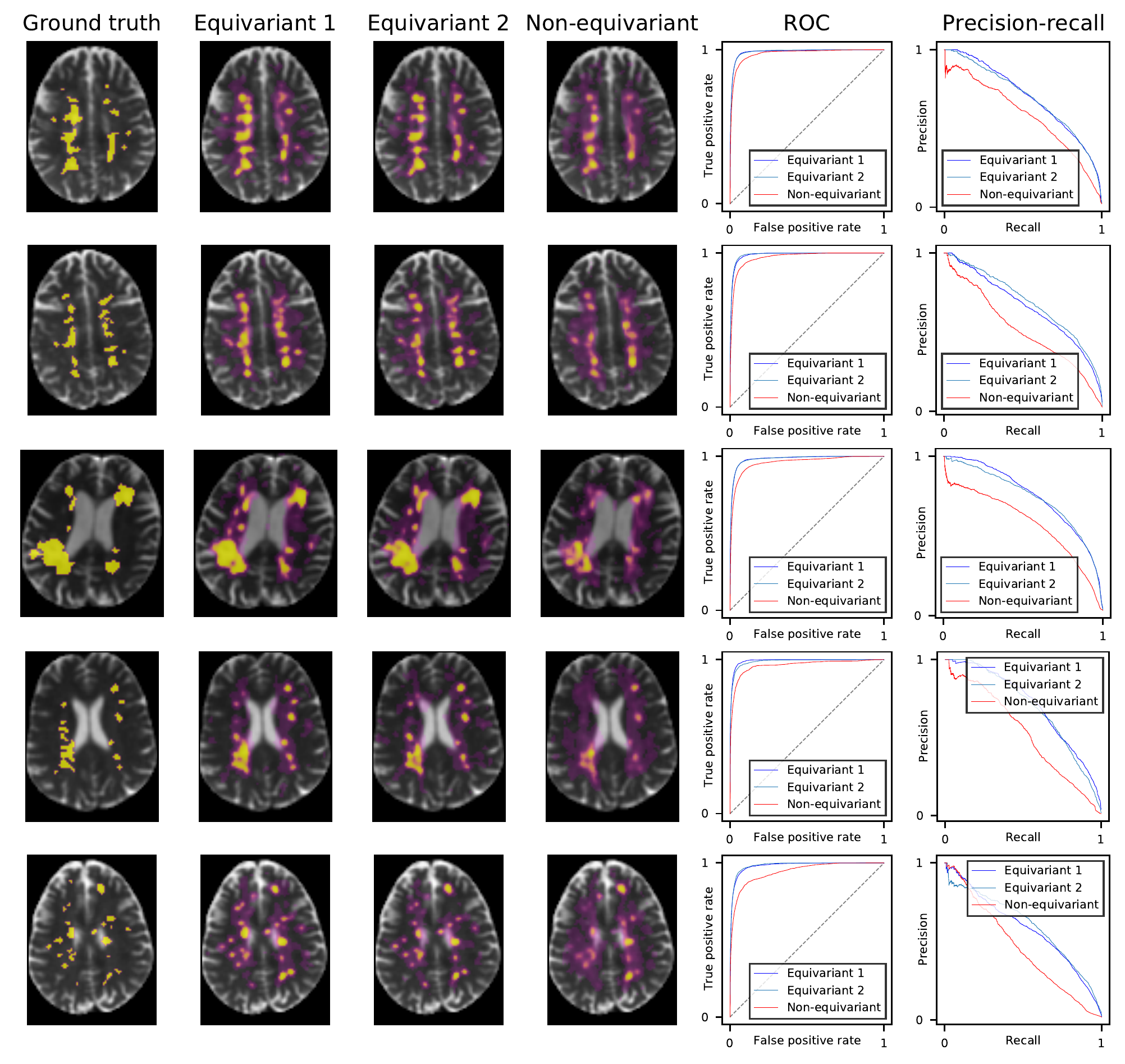}
  \caption
  {Segmentation of multiple-sclerosis lesions in five scans from the validation set. (a) Ground truth of one example slice per scan, (b) predictions for that slice using l\_TP1\_1+4 (the best equivariant model in terms of Avg-Prec and Dice score on the entire validation set), (c) predictions for that slice using l\_TP1\_1+3 (the best equivariant model in terms of AUC on the entire validation set), (d)
  predictions for that slice using the best non-rotation-equivariant model (n\_4), (e) ROC curves of all models on the full scans,
  (f) precision-recall curves of all models on the full scans.
  While the equivariant models are very certain (yellow areas) at most positions, the non-rotation-equivariant model has large areas of high uncertainty (purple areas).}
  \label{fig:slices}
\end{figure}

Figure~\ref{fig:slices} shows the segmentation of six validation samples by presenting one example slice per scan with its ground truth and the predictions from
two equivariant models, the best using \texttt{late} and the best using \texttt{gradual} $q$-reduction, and from the best non-rotation-equivariant model.
Additionally, it shows the ROC and precision-recall curves of the models.
While all models roughly predict the ground truth, the equivariant models predict it more accurately.
It can especially be seen that while the equivariant models predict the MS lesions with high confidence and have
very small values outside the areas around the lesions, the non-rotation-equivariant model is very uncertain
at many positions. This also explains the low AvgPrec scores of the non-rotation-equivariant models.

\begin{figure}[t]
    \centering
    \begin{tikzpicture}[scale=0.8]
    \begin{semilogxaxis}[
        width=1.2\textwidth,
        height=0.6\textwidth,
        xlabel={Number of parameters},
        ylabel={AvgPrec},
        ymin=0.05,
        ymax=0.65,
        xmin=40000,
        enlargelimits=0.1,
        legend style={at={(0.98,0.98)},anchor=north east},
        legend cell align={left},
    ]

        \addplot+[
            only marks,
            point meta=explicit symbolic,
            mark size=2,
            mark=x,
        ]
        coordinates {

            (446234,0.4739)
            (307136,0.5922)
            (320066,0.5911)
            (461344,0.6088)
            (465914,0.6061)
            (27511,0.4909)
            (485419,0.6015)
            (624692,0.5936)
            (485419,0.5809)
            (69671,0.5025)
            (333464,0.6118)
            (419761,0.6068)
            (470404,0.6001)
            (492242,0.6163)
            (585114,0.6198)
            (386224,0.6137)
            (626856,0.6175)
            (651020,0.6139)
            (750467,0.6112)
            (1145202,0.5901)
            (836964,0.6174)
            
             (344842,0.6112)
   			(446089,0.5918)
            (890931,0.5524)
            (503620,0.6074)
            (655748,0.6044)
            (802396,0.5898)
            (611442,0.6137)
            (1067914,0.5399)
            (633830,0.5909)
            (662398,0.6299)
            (662398,0.5949)
           (39184,0.5816)
           (39184,0.5369)
           (671887,0.6033)
           (674795,0.6171)
            (776795,0.6139)
            (848342,0.5875)
            
            (510594,0.6184)
  		    (622820,0.5649)
   			(669372,0.6095)
            (822748,0.5775)
            (855316,0.5657)
  
           (367998,0.6072)
           (271509,0.5980)
           (269449, 0.6150)
    
            (54951,0.4661)
            (61681,0.5354)
            (144031,0.5724)
            (201510,0.5719)
            (234670,0.5617)
            (200465,0.5698)
            (273611,0.5620)
            (273206,0.5265)
            (442521,0.5676)
            (220258,0.5492)
            (329640,0.5701)
            (329640,0.5484)
            (66171,0.5276)
            (66171,0.5501)
            (681385,0.5252)
            (402786,0.5568)
            (355968,0.5301)
            
            (72444,0.5772)
            (133579,0.5933)
            (359140,0.5710)
            
            (488315,0.5428)
            
            (318654,0.5814)
            
            (235820,0.4992)
            (218312,0.5002)
            (210520,0.4665)
            
        };
        \addlegendentry[color=blue]{\small Equivariant methods}

        \addplot+[
            only marks,
            point meta=explicit symbolic,
            mark size=1.5,
            mark=*,
        ]
        coordinates {
            (19590724,0.0709)
            (12720208,0.0554)
             (52909,0.3699)
            (116936,0.2691)
            (49779,0.3964)
            (622820,0.3319)
            (97899,0.4540)
            (216921,0.4532)
            (31009,0.1433)
            (64391,0.4429)
        };
        \addlegendentry[color=red]{\small Non-equivariant methods}
    \end{semilogxaxis}
\end{tikzpicture}
    \caption{
    Comparison of average precision (AvgPrec) scores of equivariant models (blue)
     and non-rotation-equivariant models (red) in relation to their number of parameters.
      The two non-rotation-equivariant models with more parameters (but similar feature map sizes)
        as the equivariant models perform much worse than the non-rotation-equivariant models with much fewer parameters.
        All shown equivariant models generalize better than
        the non-rotation-equivariant models. As many equivariant models, including the best ones, have more parameters than 
        the best non-rotation-equivariant models, the equivariance introduces a quality improvement that cannot only be explained by a reduction of the number of parameters. Instead, the equivariance allows the use of many parameters that can effectively capture the essence of the dataset so that the model does not underfit while still restricting it so that overfitting is effectively reduced without using additional regularization.
      }
    \label{fig:param_vs_AP}
\end{figure}
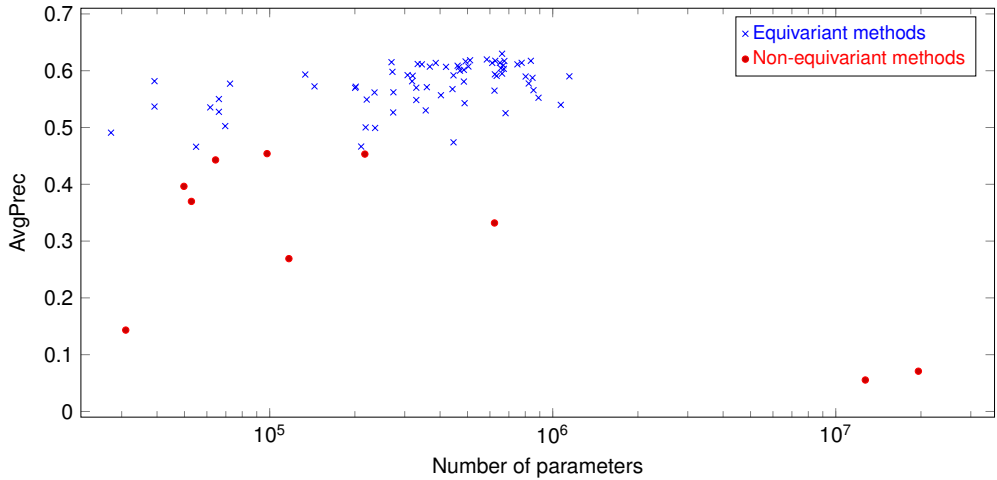
Figure~\ref{fig:param_vs_AP} compares the performance (AvgPrec) of the equivariant models and
the non-rotation-equivariant models in relation to their number of parameters.
The non-rotation-equivariant models with fewer parameters perform much better than
the non-rotation-equivariant models with more parameters but similar feature map sizes as the equivariant models.
But as many equivariant models, including the best ones, have more parameters than 
the best non-rotation-equivariant models, the superiority of the equivariant models cannot (only) be explained by
a reduction of the number of parameters. The absolute and relative differences between the training and validation
results of most equivariant models are much larger than for the non-rotation-equivariant models,
indicating that the proposed equivariant layer introduces some regularization.
But there are also effects beside regularization that enable the equivariant models to achieve better results,
as most equivariant models, including the best ones, outperform the non-rotation-equivariant models in the training metrics as well.

\paragraph{Behaviour on Reduced Training Dataset}
\begin{figure}[t]
  \centering
  \subfloat{\begin{tikzpicture}[scale=0.6]
    \begin{axis}[
        width=.55\textwidth,
        xlabel={Relative training set size},
        ylabel={ROC AUC},
        ymin=0.7,
        ymax=1.0,
        xtick={0,20,40,60,80,100},
        xticklabels={$0\%$,$20\%$,$40\%$,$60\%$,$80\%$,$100\%$},
        minor x tick num=3,
        minor y tick num=1,
        legend style={at={(0.98,0.02)},anchor=south east},
        legend cell align={left},
    ]
        \addplot+[
            point meta=explicit symbolic,
            mark=x,
            mark size=2,
        ]
        coordinates {
            (01.32,0.7282)
            (02.63,0.8796)
            (06.58,0.9006)
            (13.16,0.9246)
            (26.32,0.9545)
            (65.79,0.9719)
            (100,0.9817)
        };
        \addlegendentry[color=blue]{\small Equivariant}

        \addplot+[
            point meta=explicit symbolic,
            mark=*,
            mark size=1.5,
        ]
        coordinates {
            (01.32,0.7383)
            (02.63,0.8434)
            (06.58,0.8981)
            (13.16,0.9203)
            (26.32,0.9120)
            (65.79,0.9505)
            (100,0.9536)
        };
        \addlegendentry[color=red]{\small Non-equivariant}
    \end{axis}
\end{tikzpicture}}
  \subfloat{\begin{tikzpicture}[scale=0.6]
    \begin{axis}[
        width=.55\textwidth,
        xlabel={Relative training set size},
        ylabel={AvgPrec},
        ymin=0.0,
        ymax=0.65,
        xtick={0,20,40,60,80,100},
        xticklabels={$0\%$,$20\%$,$40\%$,$60\%$,$80\%$,$100\%$},
        minor x tick num=3,
        minor y tick num=3,
        legend style={at={(0.98,0.02)},anchor=south east},
        legend cell align={left},
    ]
        \addplot+[
            point meta=explicit symbolic,
            mark=x,
            mark size=2,
        ]
        coordinates {
            (01.32,0.0341)
            (02.63,0.1345)
            (06.58,0.1838)
            (13.16,0.3193)
            (26.32,0.4431)
            (65.79,0.5662)
            (100,0.6299)
        };
        \addlegendentry[color=blue]{\small Equivariant}

        \addplot+[
            point meta=explicit symbolic,
            mark=*,
            mark size=1.5,
        ]
        coordinates {
            (01.32,0.0219)
            (02.63,0.0938)
            (06.58,0.1897)
            (13.16,0.2167)
            (26.32,0.1609)
            (65.79,0.4304)
            (100,0.4540)
        };
        \addlegendentry[color=red]{\small Non-equivariant}
    \end{axis}
\end{tikzpicture}}
  \subfloat{\begin{tikzpicture}[scale=0.6]
    \begin{axis}[
        width=.55\textwidth,
        xlabel={Relative training set size},
        ylabel={Dice},
        ymin=0.0,
        ymax=0.65,
        xtick={0,20,40,60,80,100},
        xticklabels={$0\%$,$20\%$,$40\%$,$60\%$,$80\%$,$100\%$},
        minor x tick num=3,
        minor y tick num=3,
        legend style={at={(0.98,0.02)},anchor=south east},
        legend cell align={left},
    ]
        \addplot+[
            point meta=explicit symbolic,
            mark=x,
            mark size=2,
        ]     
        coordinates {
            (01.32,0.0184)
            (02.63,0.2182)
            (06.58,0.2575)
            (13.16,0.3806)
            (26.32,0.4701)
            (65.79,0.5572)
            (100,0.5918)
        };
        \addlegendentry[color=blue]{\small Equivariant}

        \addplot+[
            point meta=explicit symbolic,
            mark=*,
            mark size=1.5,
        ]
        coordinates {
            (01.32,0.0189)
            (02.63,0.1615)
            (06.58,0.2572)
            (13.16,0.2962)
            (26.32,0.1862)
            (65.79,0.4561)
            (100,0.4846)
        };
        \addlegendentry[color=red]{\small Non-equivariant}
    \end{axis}
\end{tikzpicture}}
  \caption
  {Comparison of the best equivariant model with the best non-rotation-equivariant model both trained on reduced subsets of the training set.
    The plots show the AUC scores (left), AvgPrec scores (middle), and Dice scores (right) of our best equivariant model, l\_TP1\_1+4,,  (blue)
    and the best non-rotation-equivariant model, n\_4, (red) trained on reduced subsets where the subset size ($x$-axis) is described relative to the full training set size. The scores are measured on the full validation set.
    Our equivariant model trained on only $26 \%$ 
    of the training scans achieves more than $100 \%$  of the AUC score, $97.6 \%$ 
    of the AvgPrec score, and $97.0 \%$ 
    of the Dice score of the non-rotation-equivariant model trained on the full dataset. When trained on $66 \%$
of the training scans, our equivariant model outperforms the non-rotation-equivariant model trained on the full training set by $1.9 \%$ in AUC score, by $24.7 \%$
in AvgPrec score, and by $15.0 \%$ in Dice score. Therefore, equivariant methods generalize better and thus require smaller training sets, as expected. Moreover, when matching the training set size for both methods, the equivariant method performs better in almost all cases, and never considerably worse.}
  \label{fig:reduced_dataset}
\end{figure}
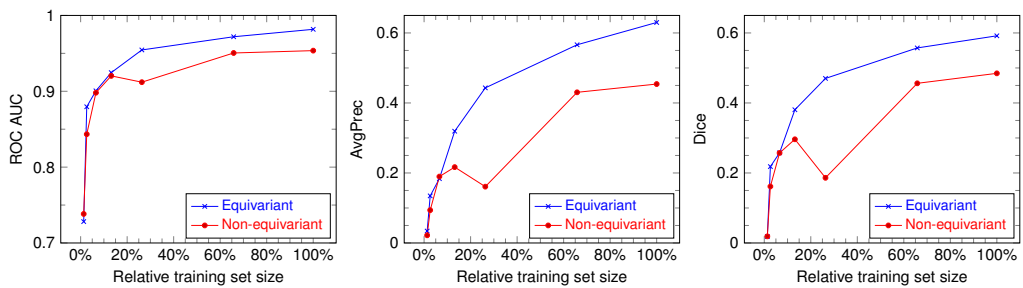
\begin{figure}[t]
  \centering
  \includegraphics[width=.9\textwidth]{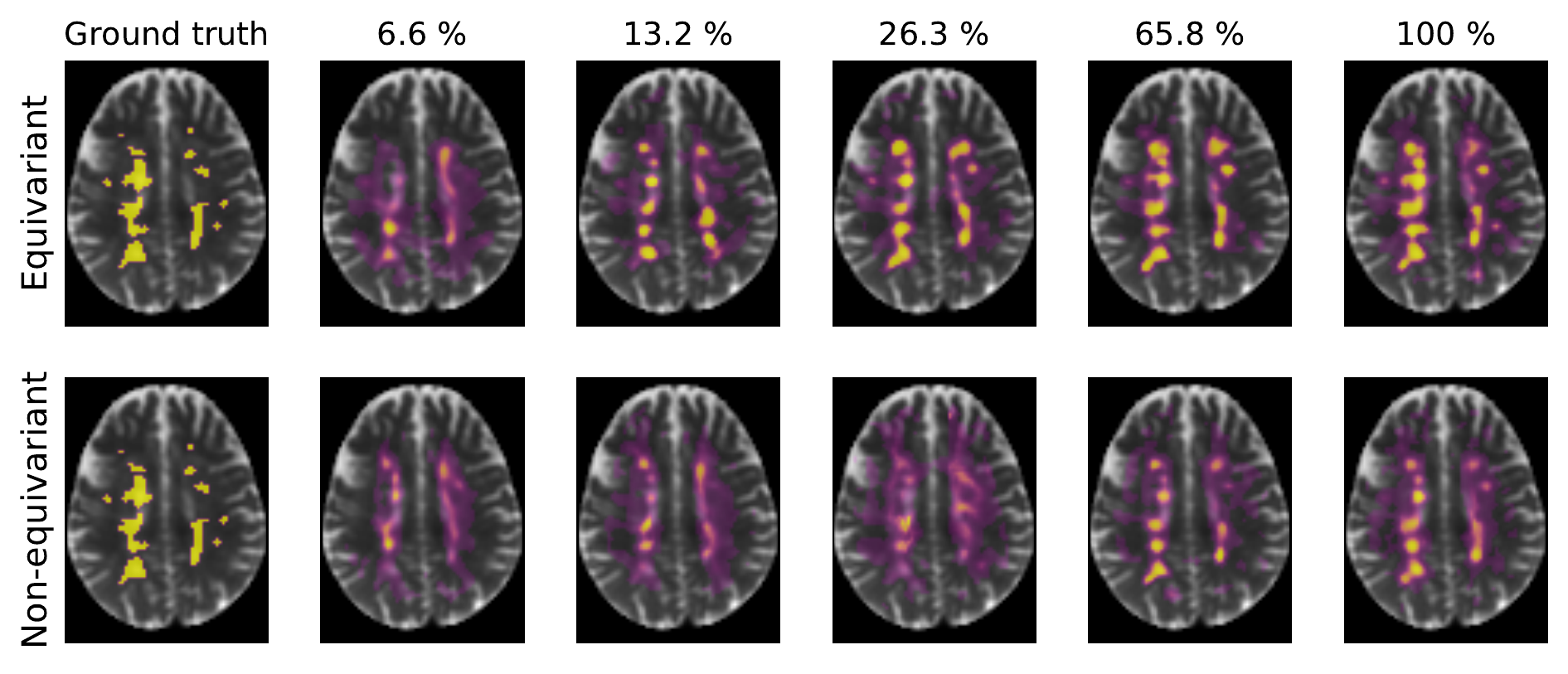}
  \caption
  {Segmentation of multiple-sclerosis lesions in one scan from the validation set using an equivariant and a non-rotation-equivariant model both trained on reduced subsets of the training set. 
  From left to right, we show the ground truth segmentation and predictions using our best equivariant model, l\_TP1\_1+4, (top) and the best non-rotation-equivariant model, n\_4, (bottom) trained on $6.6 \%$, $13.2 \%$, $26.3 \%$, $65.8 \%$, and $100 \%$ of the training scans. While the equivariant model already achieves quite accurate segmentations with $26.3 \%$ of the training samples, the segmentations of the non-rotation-equivariant model only start getting accurate with $65.8 \%$ of the training samples, which also indicates that our equivariant model generalizes faster.}
  \label{fig:slices_reduced}
\end{figure}
In order to analyze their generalization capabilities, we trained our best (measured in AvgPrec and Dice score) equivariant model and the best non-rotation-equivariant model on reduced subsets of the training set and found that the equivariant model outperforms the non-rotation-equivariant model in almost all subset sizes.
Figure~\ref{fig:reduced_dataset} shows the AUC, AvgPrec, and Dice scores of both models on different subset sizes.
Our model trained on only $26\%$
    of the training scans achieves $100.1 \%$
    of the AUC score, $97.6 \%$ 
    of the AvgPrec score, and $97.0 \%$ 
    of the Dice score of the non-rotation-equivariant model trained on the \emph{full} training dataset.
When trained on $66\%$
of the training scans, our equivariant model outperforms the non-rotation-equivariant model trained on the full training set by $1.9 \%$ in AUC score, by $24.7 \%$
in AvgPrec score, and by $15.0 \%$ in Dice score.
This enables to use smaller datasets while achieving the same or better performance.
Moreover, when matching the training set size for both methods, the equivariant method performs better in almost all cases, and never considerably worse.
Figure~\ref{fig:slices_reduced} confirms these results by comparing the segmentation of one example slice using the equivariant and the non-rotation-equivariant models trained on different subset sizes. 

\paragraph{Comparison of $q$-Reduction Strategies, Basis Filters, Layers, Channel Setups, and Radial Basis Functions}
Models using \texttt{late} $q$-reduction perform much better than models using \texttt{gradual} $q$-reduction.
The \texttt{TP$\pm 1$} basis outperforms all other proposed filter bases.
For \texttt{late} $q$-reduction models, the best results are achieved using one $pq$-layer and three or four $p$-layers. For \texttt{gradual} $q$-reduction models, one $pq$-layer with either one or two $p$-layers works best.
Small changes to the number of channels do not influence the results much.
Also using order $l=2$ and $l=3$ channels neither leads to much better nor much worse performance.
The best results could be achieved with a radial basis that consists of a FC network applied to cosine radial basis functions (cosine+fc). While using
Gaussian+fc also works quite well, models using no FC network in the radial basis (cosine, Gaussian) perform much worse.

\paragraph{Training Times and Memory Consumption}
The equivariant models trained about $0.5-2.5$ days until convergence 
and required about $12-24$ GB of GPU RAM while most non-rotation-equivariant models trained only a few hours 
until convergence and only required
$<2$ GB 
of GPU memory.

The long training times of the equivariant models are caused by
\begin{inparaenum}[i)]
    \item more epochs until convergence,
    \item computation of the kernel after each weight update, and
    \item longer backpropagation chains to the kernel parameters (because the kernel does not consist of the parameters but is computed based on functions of the parameters).
\end{inparaenum} The first reason seems to have less relevance,
as models with smaller feature maps but more epochs until convergence were faster
than models with larger feature maps but fewer epochs. As the kernels in the $pq$-layers are much larger than in the $p$-layers,
the training times mostly depend on the number of $pq$-layers and their number of channels.

The GPU RAM consumption was mainly caused by the feature maps stored during the forward pass
and their gradients computed during the backward pass, which is why the models with \texttt{late} $q$-reduction
required much more memory (up to $24$ GB) than the models with \texttt{gradual} $q$-reduction (up to $16$ GB).
The difference in memory consumption between the equivariant and the reference models
is caused by \begin{inparaenum}[i)]
    \item the precomputed parts of the kernels,
    \item the intermediate values stored during kernel computation to be used in the backward pass, and
    \item the gated nonlinearities requiring to store more intermediate feature maps.
\end{inparaenum}

\section{Conclusions}\label{sec:conclusion}
In this work, we proposed $\SE(3)$-equivariant deep learning for diffusion MRI data and showed that it yields better results and decreases the required number of training samples.

The superiority of the equivariant approach over non-rotation-equivariant models may be explained by the restrictions it imposes on the filters.
As the non-rotation-equivariant models use unrestricted convolutional kernels, they need to learn the rotational equivariance from data.
This means that their parameters need to capture the equivariance and all other information contained in the dataset, whereas the parameters of the equivariant models do not need to capture the equivariance as it is already imposed by the model itself. Thus the parameters
of equivariant models can be used more effectively for capturing all other aspects of the data. This prevents underfitting and simplifies training.
Another drawback of the non-rotation-equivariant models is that they may overfit on specific orientations in the training set,
while the rotational equivariance of the proposed models prevents this.
In the validation set, the non-rotation-equivariant models cannot recognize orientations of patterns not seen during training,
while the equivariant models recognize these patterns and thus achieve better results.

This can also explain why the equivariant models require a smaller number of training samples.
When trained on only very few samples, this benefit might become less relevant compared to overfitting
on other properties of the patterns not related to rotational equivariance,
explaining why then the equivaviant and non-rotation-equivariant models perform similarly on tiny training sets,
e.g.\ the patterns in different scans may also differ in scale or
brightness,
they may be deformed or some types of patterns may only be present in some samples.

There may be several reasons why the \texttt{late} $q$-reduction strategy outperforms the \texttt{gradual} one:
Using a different $q$-sampling scheme in the output of a layer than in the input may require interpolations that may be hard to learn for the layer.
Additionally, the \texttt{late} $q$-reduction layer, \texttt{q-length weighted average}, operates point-wise in $p$-space, whereas
in \texttt{gradual} $q$-reduction, a filtering layer with a $p$-space kernel size greater than one is used, which may be harder to train.
In order to achieve the same total receptive field, \texttt{gradual} $q$-reduction requires less layers than \texttt{late} $q$-reduction, because of the larger kernel size of its $q$-reduction layer compared to the point-wise \texttt{late} $q$-reduction layer. This can explain why the best \texttt{late} $q$-reduction models have more $p$-layers than the best \texttt{gradual} models.

The reason for the \texttt{TP$\pm 1$} filter basis performing best may be that it can access more aspects of the structural dependencies between $p$- and $q$-space than the other proposed bases, as explained in Section~\ref{sec:filter_basis_options}. Another reason of its success may be that it has more capacity, noticeable by the larger number of parameters. 
From the results it may also be concluded that fine angular details detected by higher-order filters
do not seem to have much relevance, as using higher-order filters does not lead to better performance.
The benefit of using fully connected (FC) networks with the radial basis functions may be explained by their larger capacity, which allows them to better detect radial patterns.

In general, the equivariance of the proposed layer allows the use of many parameters that can effectively capture the essence of the dataset, so that the model does not underfit while still restricting it so that overfitting is effectively reduced without using additional regularization like data augmentation.
Our results show that using the proposed equivariant layer can help increasing the performance of predictions on dMRI scans, thus future research in this direction seems promising.
Besides applying the proposed equivariant network architecture to different datasets and tasks, also different architectures including the proposed layer may be developed, e.g.\ architectures with equivariant $p$-space pooling. As the proposed layer supports vector and tensor inputs and outputs,
also architectures predicting vectors, e.g.\ for fiber detection, or with diffusion tensors as input or outputs may be built (see Appendix~\ref{appendix:higher_order_outputs}).

While the derivation of the layers is mathematically involved, our public implementation can be easily used out of the box without understanding the mathematical background, and gives immediate access to the benefits of equivariant deep learning for diffusion MRI. Merely general practices are advisable such as tuning the learning rate.

\subsection*{Acknowledgments}
This work was supported by the Munich Center for Machine Learning (Grant No. 01IS18036B) and the BMBF project MLwin.

\bibliographystyle{iclr2021_conference}
\bibliography{ms}

\clearpage
\appendix
\section{Theoretical Background}\label{appendix:theo_background}
\subsection{Groups, Group Representations, and Equivariance}\label{appendix:intro_group}
Intuitively, a \emph{group}~$G$ may be seen as a
finite or infinite set of invertible and composable operations acting on some set $\mathcal{X}$.
Examples of groups are the translations in 3D, the rotations in 3D, denoted by~$\SO(3)$, and the roto-translations in 3D, denoted by~$\SE(3)$.
For a detailed formal introduction to groups, see
for example~\citet{group_1,group_3,group_6}.

If~$\mathcal{X}$ is a vector space, then a group~$G$ can act on~$\mathcal{X}$ through a \emph{(group) representation}~$\bm{D}^\mathcal{X}$, which is defined as a function from~$G$ to the set of invertible linear transformations (or invertible square matrices)
on $\mathcal{X}$ with the following property~\citep{group_2,group_4,TFN}:

\begin{align}
\bm{D}^\mathcal{X}_g \circ \bm{D}^\mathcal{X}_h = \bm{D}^\mathcal{X}_{g \cdot h} \qquad \forall g, h \in G,
\end{align}
where $\bm{D}^\mathcal{X}_g$ denotes the representation $\bm{D}^\mathcal{X}$ applied to $g \in G$ and is called a representation of the group element $g$ on $\mathcal{X}$, the operator ``$\circ$'' denotes function composition, and ``$\cdot$'' denotes composition of group elements (group multiplication).

Now consider a function $f \colon \mathcal{X} \to \mathcal{Y}$ mapping between the vector spaces~$\mathcal{X}$ and~$\mathcal{Y}$.
Given a group $G$ and two representations~$\bm{D}^\mathcal{X}$,~$\bm{D}^\mathcal{Y}$ of $G$ acting
on~$\mathcal{X}$ and~$\mathcal{Y}$, respectively, the function $f$ is called \emph{equivariant under~$G$}
if the following holds~\citep{group_2,group_4,TFN}:

\begin{align}\label{eq:def_equi_repr}
  f\left(\bm{D}^\mathcal{X}_g [\bm{x}]\right) = \bm{D}^\mathcal{Y}_g \left[f(\bm{x})\right] \qquad \forall g \in G \qquad \forall \bm{x} \in \mathcal{X}.
\end{align}
Intuitively, equivariance means that if the input of $f$ is transformed by some group element $g$ then this leads to the same results as applying $f$ to the non-transformed input and transforming the output of $f$ by $g$.
Thus, rotation-equivariant neural networks produce the same segmentation results
(up to rotation) regardless of how the input image is rotated.

If $f$ is equivariant and $\bm{D}^\mathcal{Y}_g$ is the identity for all $g \in G$, then
$f$ is said to be \emph{invariant under $G$}~\citep{group_4,TFN}.
Invariance means that the output of $f$ does not change if its input is transformed by some $g \in G$.
Thus, rotation-equivariant neural networks produce the same image-classification results (not rotatable) regardless of how the input image is rotated.

For a group $G$ there may be multiple different representations on the same
vector space $\mathcal{X}$~\citep{group_4,group_5}.
A representation $\bm{D}^\mathcal{X}$ of $G$ on $\mathcal{X}$ is called \emph{reducible} if there exists a basis transformation in $\mathcal{X}$,
described by the invertible square-matrix $\bm{Q}$, such that the representation in the new basis, $\bm{Q}^{-1}\bm{D}^\mathcal{X}_g\bm{Q}$, is a block-diagonal matrix, i.e.~whose blocks $\bm{D}^{\mathcal{X}_i}_g$ act on subspaces $\mathcal{X}_i$, with the conditions that $\mathcal{X}$ is the direct sum of all $\mathcal{X}_i$ and that each $\mathcal{X}_i$ is closed under the corresponding representation $\bm{D}^{\mathcal{X}_i}_g$, i.e.~\(
\bm{D}^{\mathcal{X}_i}_g\bm{x}_i \in \mathcal{X}_i \; \forall g \in G \; \forall \bm{x}_i \in \mathcal{X}_i.
\)
The intuition behind $\bm{Q}^{-1}\bm{D}^\mathcal{X}_g\bm{Q}$ is that the linear operator $\bm{D}^\mathcal{X}_g$ is applied in a different basis, i.e.~$\bm{Q}^{-1}\bm{D}^\mathcal{X}_g\bm{Q}y$ means: map $y$ to basis of $\bm{D}^\mathcal{X}_g$ using $\bm{Q}$, apply $\bm{D}^\mathcal{X}_g$, map it back to basis of $y$ using $\bm{Q}^{-1}$.
A representation that cannot be decomposed in such a way is called an \emph{irreducible} representation or \emph{irrep}.

\subsection{Spherical Tensors and Related Concepts}\label{appendix:intro_tensor}
\emph{Tensors}
are quantities transforming under rotations in a very specific way depending on their order $l \in \mathbb{N}_0$, and which in 3D
can be described using $3^l$ components, which in the case of real tensors are real numbers. 
For the purposes of this work (i.e.~considering only contravariant, real 3D tensors), 
a tensor $\bm{T}$ of order $l$ transforms under the rotation $g \in \SO(3)$ into the tensor $\bm{T}'$ (of same order $l$) as follows~\citep{tensor_1,tensor_2}:

\begin{align}\label{eq:tensor_rotation_property}
  T_{\mu_1, \mu_2 \dots \mu_l}' = \sum_{\nu_1, \nu_2 \dots \nu_l}
  (\mathcal{R}_g)_{\mu_1, \nu_1} (\mathcal{R}_g)_{\mu_2, \nu_2} \dots (\mathcal{R}_g)_{\mu_l, \nu_l}
  T_{\nu_1, \nu_2 \dots \nu_l},
\end{align}
where $\mu_1, \mu_2 \dots \mu_l, \nu_1, \nu_2 \dots \nu_l$ are indices with values $1, 2, 3$ used to access the components of the tensors and
$\bm{\mathcal{R}}_g$ is the rotation matrix of $g$.
\emph{Scalars} are tensors of order $l=0$ and do not transform under rotations, whereas \emph{vectors} are tensors of order $l=1$.

\subsubsection{Spherical Tensors, Spherical Harmonics, and Wigner D-Matrices}\label{appendix:spherical_tens_harm_D}
The tensors described so far are called \emph{Cartesian tensors}.
Although these tensors obey the simple transformation rule of Eq.~\eqref{eq:tensor_rotation_property},
they may be \emph{reducible} w.r.t.~rotations, meaning that they can be decomposed into other possibly lower-order tensors, whose components can be computed as linear combinations of the original tensor components, and the resulting tensors can be rotated independently~\citep{tensor_3,tensor_4}.
Tensors can instead be represented in a different basis, the \emph{spherical basis}, in which they are always irreducible.
Such tensors are called \emph{spherical tensors} and their components are called \emph{spherical components}~\citep{tensor_1,tensor_3,tensor_4,tensor_5,tensor_7}.
Under the rotation $g \in \SO(3)$, a spherical tensor $\bm{\mathcal{T}}^{(l)}$ of order $l$ transforms into
$\bm{\mathcal{T}}'^{(l)}$ using the irreducible representations of $\SO(3)$ as follows~\citep{tensor_10}:

\begin{align}\label{def:real_spherical_tensor}
  \bm{\mathcal{T}}'^{(l)} = \bm{D}^{(l)}_{g^{-1}}\bm{\mathcal{T}}^{(l)},
\end{align}
where $\bm{D}^{(l)}_{g^-1}$, called (real) \emph{Wigner D-matrix of order $l$}, is an irreducible representation of $g^{-1}$ and is orthogonal.

Spherical tensors of order $l$ can be described using $2l+1$ either complex or real components.
Describing them using the same number of either complex or real components is possible
due to symmetries of spherical tensors~\citep{tensor_4,tensor_5,tensor_10}.
As real numbers require less memory for storage and are more efficient from a computational point of view~\citep{tensor_5},
we prefer to use real components and denote the vector space of (the real-valued description of) order $l$ spherical tensors by $\mathcal{S}^{(l)}$.

For the cases of $l=0$ (scalars) and $l=1$ (vectors), the (real) Wigner D-matrices are (up to a reordering
of its components in the case of $l=1$) given by $D^{(0)}_g = 1$ and $\bm{D}^{(1)}_g = \bm{\mathcal{R}}_g$~\citep{tensor_10,tensor_5,tensor_7,TFN}. 

There is a special set of functions mapping points on the sphere $S^2$ to spherical tensors,
the \emph{(real) spherical harmonics} $\bm{Y}^{(l)}$, where $l$ denotes the order of the spherical harmonics and is equal to the order of its outputs.

Under a rotation $g \in \SO(3)$, the (real) spherical harmonics transform like spherical tensors using the
(real) Wigner D-matrices~\citep{tensor_10,tensor_7,TFN}:

\begin{align}\label{eq:real_spherical_harmonics_rotation_1}
  \left(g \bm{Y}^{(l)}\right)(\bm{n}) = \bm{Y}^{(l)}\left(\bm{\mathcal{R}}_{g^{-1}}\bm{n}\right)
  = \bm{D}^{(l)}_{g^{-1}}\left[ \bm{Y}^{(l)}(\bm{n})\right],
\end{align}
where $\bm{n}$ is a unit vector in $\mathbb{R}^3$ representing a point on the sphere, meaning $\bm{n} \in S^2$.

For the case $l=0$ it holds that $\bm{Y}^{(0)}(\bm{n}) = \mathrm{const.}$, while for $l=1$ it holds (up to a reordering of the components of $\bm{n}$) that $\bm{Y}^{(1)}(\bm{n}) = c\bm{n}$ with some constant factor $c \in \mathbb{R}$~\citep{tensor_10,TFN}.

\subsubsection{Tensor Product and Clebsch--Gordan Coefficients}\label{appendix:CG}
Two spherical tensors
$\bm{\mathcal{T}}^{(l_1)}$, $\bm{\mathcal{U}}^{(l_2)}$ of orders $l_1, l_2$
can be coupled using the \emph{tensor product}, denoted by $\otimes$,
which for spherical tensors results in the following direct sum (concatenation) of spherical tensors~\citep{tensor_2,tensor_3,tensor_4,tensor_6,tensor_7,TFN,NBody}:
\begin{align}\label{eq:TP_direct_sum}
  \bm{\mathcal{T}}^{(l_1)} \otimes \bm{\mathcal{U}}^{(l_2)} \coloneqq \bigoplus_{l = |l_1 - l_2 |}^{l_1 + l_2}
  \left(\bm{\mathcal{T}}^{(l_1)} \otimes \bm{\mathcal{U}}^{(l_2)}\right)^{(l)},
\end{align}
where $\bigoplus$ denotes the direct sum, and with
\begin{align}\label{eq:TP_spherical}
  \left(\bm{\mathcal{T}}^{(l_1)} \otimes \bm{\mathcal{U}}^{(l_2)}\right)^{(l)}_m = \sum_{m_1=-l_1}^{l_1}\sum_{m_2=-l_2}^{l_2}
  C^{(l, m)}_{(l_1, m_1)(l_2, m_2)} \mathcal{T}^{(l_1)}_{m_1} \mathcal{U}^{(l_2)}_{m_2},
\end{align}
where $C^{(l, m)}_{(l_1, m_1)(l_2, m_2)}$ are predefined scalars, the so-called (real) \emph{Clebsch--Gordan coefficients}, $l$ is the order of the resulting tensor, and $m$ with $-l \le m \le l$ is used to index its components. 
The meaningful orders $l$ of tensors resulting from the tensor product follow from the fact that the Clebsch--Gordan coefficients are only non-zero if the following holds:
\begin{align}\label{eq:CG_non_zero}
  |l_1 - l_2 | \le l \le l_1 + l_2,
\end{align}
and as such the tensors of other orders are zero.
Additionally, it needs to hold that $m = m_1 + m_2$ for the Clebsch--Gordan coefficients to be non-zero.
As the values of the spherical harmonics are spherical tensors, spherical harmonics can be coupled with spherical tensors via a tensor product in the same way.

An important property of the tensor product between two spherical tensors
is that it is equivariant
under rotations, meaning that the resulting spherical tensors, as defined in Eq.~\eqref{eq:TP_spherical},
transform as follows~\citep{tensor_5,TFN}:
\begin{align}\label{eq:TP_equivariance}
  \left(\bm{D}^{(l_1)}_g\bm{\mathcal{P}}^{(l_1)} \otimes \bm{D}^{(l_2)}_g\bm{\mathcal{Q}}^{(l_2)}\right)^{(l)}
  = \bm{D}^{(l)}_g\left(\bm{\mathcal{P}}^{(l_1)} \otimes \bm{\mathcal{Q}}^{(l_2)}\right)^{(l)} \qquad \forall g \in \SO(3),
\end{align}
    where $\bm{D}$ are the Wigner D-matrices.
For some special values of $(l_1, l_2, l)$, the tensor product of spherical tensors is related to some well-known operations:
The case of $(0, 0, 0)$ is equal to normal multiplication of two scalars, $(0, 1, 1)$ and $(1, 0, 1)$
are equal to scalar multiplication of a vector, while the cases $(1, 1, 0)$ and $(1, 1, 1)$
are proportional to the dot product and the cross product of two vectors, respectively~\citep{tensor_2,tensor_6,tensor_13,TFN}.

\subsubsection{Multi-Channel Spherical Tensors and Spherical-Tensor Fields}\label{appendix:tensor_field}
So far spherical tensors have been defined with a single order $l$.
Following~\citet{NBody}, multiple such spherical tensors can be combined using the direct sum (concatenation), denoted by $\bigoplus$.
We will call such objects \emph{multi-channel spherical tensors} and call the spherical tensors comprising it \emph{channels}. This is related to the channels in convolutional neural networks, where every pixel/voxel has scalar channels, but the concept of channels is generalized from scalars to other tensors as e.g.\ done in~\citet{TFN,NBody}.
The type of the multi-channel spherical tensor is defined by the tuple $\bm{\tau} = (\tau_0, \tau_1, \dots)$ where
each $\tau_l \in \mathbb{N}_0$ defines the number of channels of order $l$.
The space of multi-channel spherical tensors of type $\bm{\tau}$ thus is defined
as
\(
  \mathcal{S}^{\bm{\tau}} := \bigoplus_{l=0}^\infty \bigoplus_{c^{(l)} = 1}^{\tau_l} \mathcal{S}^{(l)}
\),
where $c^{(l)}$ is used to index the channels of order $l$, and $\mathcal{S}^{(l)}$ is the vector space of spherical tensors of order $l$.
The total number $C$ of channels is $C = \sum_{l=0}^\infty \tau_l$.
In this work, the channel is often directly indexed using $c \in \{1, \dots, C\}$ and the order of this channel may be denoted by $l(c)$.

A 3D \emph{(multi-channel) spherical-tensor field}~$\bm{I}$ of type $\bm{\tau}$ assigns a type-$\bm{\tau}$ spherical tensor to every position in space, i.e.~$\bm{I} \colon \mathbb{R}^3 \to \mathcal{S}^{\bm{\tau}}$.
Such a tensor field has special transformation properties under rotations and translations~\citep{tensor_1,tensor_5,3dSteerableCNNs}:
\begin{align}
   (g, \mathcal{T}_{\bm{t}})[\bm{I}](\bm{x}) = \bm{D}^{\bm{\tau}}_g\bm{I}\left(\mathcal{\bm{R}}_{g^{-1}}(\bm{x}-\bm{t})\right) \qquad \forall \bm{x} \in \mathbb{R}^3,
\end{align}
where $(g, \mathcal{T}_{\bm{t}}) \in \SE(3)$ is a roto-translation with rotation $g \in \SO(3)$ followed by translation $\bm{t} \in \mathbb{R}^3$, with the multi-channel Wigner D-matrix $\bm{D}^{\bm{\tau}}_g$ of type $\bm{\tau}$, and with the rotation matrix $\mathcal{\bm{R}}_{g^{-1}}$ of the inverse rotation $g^{-1}$.
This means that the value for position $\bm{x}$ of the transformed tensor field is read from
the original position (before the transformation) given by $\mathcal{\bm{R}}_{g^{-1}}(\bm{x}-\bm{t})$.
Additionally, the resulting spherical tensor is transformed according to the rotation $g$ using $\bm{D}^{\bm{\tau}}_g$.
Note that the latter transformation does not depend on the translation.
The matrix $\bm{D}^{\bm{\tau}}_g$ can be built as direct sum of Wigner D-matrices $\bm{D}^{(l)}_g$, i.e.~$\bm{D}^{\bm{\tau}}_g := \bigoplus_{l=0}^\infty \bigoplus_{c^{(l)} = 1}^{\tau_l} \bm{D}^{(l)}_g$. 
This means that each channel transforms independently but the tensor components within each channel influence each other during transformation.
If for example the tensor field represents a 3D MR image with three contrasts ($T_1$, $T_2$, proton density), then it would contain three scalar channels (the contrasts)
but no vector channels, as each of the channels transforms independently whereas the three parts
of a vector channel would influence each other during transformation because they jointly express directions in the 3D image space~\citep{3dSteerableCNNs,NBody}.

\section{Implementation Details}\label{appendix:implementation}

\subsection{Radial Basis Functions}\label{appendix:radial_basis_fn}
\subsubsection{Gaussian Radial Basis Functions as used in~\citet{3dSteerableCNNs}}
A set of Gaussian radial basis functions can be defined as
\begin{align}\label{eq:gauss_radial_fn}
      \varphi_{\text{Gaussian}}^{(k)}(x) \coloneqq \text{exp}\left(-\frac{(x-\mu_k)^2}{2 \sigma^2}\right)
\end{align}
with (predefined or learned) means $\mu_k$ and (predefined or learned) variance $\sigma^2$, where $k$ identifies each function in the set and has values from $1$ to the radial basis size $K$.

\subsubsection{Cosine Radial Basis Functions as used in~\citet{e3nn}} 
A set of cosine radial basis functions can be defined as
\begin{align}\label{eq:cosine_radial_fn}
       \varphi_{\text{cos}}^{(k)}(x) \coloneqq
   \begin{cases}
        \cos^2 (\gamma (x - \mu_{k}) \frac{\pi}{2}) & 1 \geq \gamma (x - \mu_{k}) \geq -1 \\
        0 & \mathrm{otherwise}
    \end{cases}
\end{align}
with (predefined or learned) reference points $\mu_k$ and (predefined) normalization factor $\gamma$,
where $k$ identifies each function in the set and has values from $1$ to the radial basis size $K$.

\subsubsection{Fully Connected Neural Network applied to Radial Basis Functions as used in~\citet{TFN,e3nn}}
First some set of radial basis functions $\varphi'^{(i)}$ is applied to the input $x$. The the output of each radial basis function in this set is then treated as input neuron to a fully connected neural network.
The radial basis size $K$
defines the number of neurons of the output layer of that network, while the number of hidden layers and neurons in these layers are hyperparameters.
In the case of a two-layer network, the radial basis function is defined as:
\begin{align}
  \varphi_{\text{NN}}^{(k)}(x) \coloneqq b_{k}^{(2)} + \sum_{j}W_{k, j}^{(2)}\text{ReLU}\left(b_{j}^{(1)} + \sum_{i}W_{j, i}^{(1)}\varphi'^{(i)}(x)\right),
\end{align}
where $\bm{W}^{(2)}$, $\bm{W}^{(1)}$, $\bm{b}^{(2)}$, and $\bm{b}^{(1)}$ are learned parameters, $j$ and $i$ are indices of the hidden respectively input layer,
 and $\varphi'^{(i)}$ is some set of radial basis functions.

\subsection{Equivariant Nonlinearities}\label{appendix:theo_non_lin}
When using spherical-tensor fields (see Appendix~\ref{appendix:tensor_field}) as feature maps like used in the layer proposed in Eq.~\eqref{eq:conv_layer_definition},
elementwise nonlinearities like ReLU are in general not equivariant under rotations~\citep{3dSteerableCNNs}.
Instead special nonlinearities like \emph{tensor product nonlinearities}~\citep{NBody},
\emph{norm-nonlinearities}~\citep{harmonicNetworks}, \emph{squashing nonlinearities}~\citep{dynamic_routing_capsules},
or \emph{gated nonlinearities}~\citep{3dSteerableCNNs} are required.

Note that as scalars are invariant under rotations, elementwise nonlinearities (like ReLU) can be used for $l=0$ channels~\citep{3dSteerableCNNs}.

\subsection{Discretization}\label{appendix:impl_discretization}
\paragraph{Discretized Feature Maps}
The layer definition~\eqref{eq:conv_layer_definition} assumes feature maps over the space $\mathbb{R}^3 \oplus \mathbb{R}^3$ and these feature maps would require infinite memory.
Thus, in practice, feature maps are discretized using sampling schemes.
In this work, the sampling schemes used in dMRI scans are also used for the feature maps,
so the $p$-space is sampled on a finite Cartesian 3D grid, while the $q$-space uses a predefined, not necessarrily regular, finite sampling scheme.

This means that the discretized $p$-space has the form:
\begin{align}
    \mathcal{P} \coloneqq \mathcal{P}_x \times \mathcal{P}_y \times \mathcal{P}_z \subset \mathbb{Z}^3 \subset \mathbb{R}^3,
\end{align}
with $\mathcal{P}_x = \{1, \dots, P_x\}$, $\mathcal{P}_y = \{1, \dots, P_y\}$, and $\mathcal{P}_z = \{1, \dots, P_z\}$,
where $P_x, P_y, P_z$ are the sizes of the $p$-space voxel grid of the feature map.

The discretized $q$-space is the finite set
\begin{align}
    \mathcal{Q} \coloneqq \{\bm{q}_n\}_{n=1}^{Q} \subset \mathbb{R}^3,
\end{align}
consisting of the $Q$ predefined $q$-vectors $\bm{q}_n \in \mathbb{R}^3$
where $Q$ and the $\bm{q}_n$ may be different for feature maps of different layers.
While $\mathcal{Q}$ for the input, denoted by $\mathcal{Q}_\mathrm{in}$, is dictated by the input data structure,
e.g.\ the output of the previous layer or the input to the network,
$\mathcal{Q}$ for the output, denoted by $\mathcal{Q}_\mathrm{out}$, is a freely choosable hyperparameter.
This means that we can choose $Q_\mathrm{out}=1$ to ``collapse'' $q$-space in some layer. 
This may for example be used for image segmentation (one prediction for each $p$-space coordinate) or image classification (where $p$-space gets ``collapsed'' as well, namely through pooling).
Note that $\mathcal{Q}_\mathrm{out} = {(0,0,0)}$ is required to achieve equivariance under rotations in $q$-space, as the $q$-space coordinate offsets used in the angular and radial basis are not rotation equviariant for other $\mathcal{Q}_\mathrm{out}$. For invariance under rotations in $q$-space, only scalar ($\lo=0$) output channels must be used, at least in the final layer.

Using these definitions, a discretized feature map $\hat{\bm{I}}$ of type $\bm{\tau}$ can be defined as
\begin{align}
    \hat{\bm{I}} \colon \mathcal{P} \oplus \mathcal{Q} \to \mathcal{S}^{\bm{\tau}},
\end{align}
which can be represented using an array of size
\begin{align}
  \dim\left(\mathcal{S}^{\bm{\tau}}\right) \times P_z \times P_y  \times P_x \times Q.
\end{align}

\paragraph{Discretized Convolutional Layer}
A layer $\hat{\mathcal{L}}$ working on these discretized feature maps is a function of the form
\begin{align}
  \hat{\mathcal{L}}\colon (\mathcal{P}_\mathrm{in} \oplus \mathcal{Q}_\mathrm{in} \to \mathcal{S}^{\taui}) &\to
   (\mathcal{P}_\mathrm{out} \oplus \mathcal{Q}_\mathrm{out} \to \mathcal{S}^{\tauo}),
\end{align}
where $\mathcal{P}_\mathrm{in}$, $\mathcal{Q}_\mathrm{in}$, and $\mathcal{S}^{\taui}$ are the $p$-space, $q$-space, and spherical tensor space of the input feature map, respectively, and $\mathcal{P}_\mathrm{out}$, $\mathcal{Q}_\mathrm{out}$, and $\mathcal{S}^{\tauo}$ those of the output feature map.

This layer $\hat{\mathcal{L}}$ can be defined analogously to $\mathcal{L}$~\eqref{eq:conv_layer_definition}
by replacing the sum of $\pin$ and $\qi$ over $\mathbb{R}^3$ by sums over $\mathcal{P}_\mathrm{in}$,
the discretized input $p$-space, and $\mathcal{Q}_\mathrm{in}$, the discretized input $q$-space:
\begin{align}\label{eq:def_discretized_layer_1}
\begin{split}
\hat{\mathcal{L}}_{\mo}^{(\co)}[\hat{\bm{I}}](\po, \bm{q}_{n_\mathrm{out}})
  \coloneqq & \sum_{\ci, \cf, k}W_{\ci, \cf, \co, k}
    \sum_{\substack{\mf \in \{-\lf, \dots, \lf \}, \\ \mi \in \{-\li, \dots, \li \}}}
     C_{(\lf, \mf)(\li, \mi)}^{(\lo, \mo)}\\
     & \qquad \times \sum_{\substack{\pin \in \mathcal{P}_\mathrm{in}, \\ \bm{q}_{n_\mathrm{in}} \in \mathcal{Q}_\mathrm{in}}}
     F_{\mf}^{(\cf, k)}(\po - \pin, \bm{q}_{n_\mathrm{out}}, \bm{q}_{n_\mathrm{in}})
     \hat{I}_{\mi}^{(\ci)}(\pin, \bm{q}_{n_\mathrm{in}}).
\end{split}
\end{align}

Eq.~\eqref{eq:def_discretized_layer_1} can be rewritten as follows:
\begin{align}\label{eq:def_discretized_layer_2}
\begin{split}
  \hat{\mathcal{L}}_{\mo}^{(\co)}[\hat{\bm{I}}](\po, \bm{q}_{n_\mathrm{out}}) = & \sum_{\ci, \cf, k}W_{\ci, \cf, \co, k}
       \sum_{\substack{\mf \in \{-\lf, \dots, \lf \}, \\ \mi \in \{-\li, \dots, \li \}}}
       C_{(\lf, \mf)(\li, \mi)}^{(\lo, \mo)}\\
       & \qquad \times \sum_{\substack{\bm{p}_\mathrm{filter} \in \mathcal{P}_\mathrm{filter}, \\ \bm{q}_{n_\mathrm{in}} \in \mathcal{Q}_\mathrm{in}}}
        F_{\mf}^{(\cf, k)}(-\bm{p}_\mathrm{filter}, \bm{q}_{n_\mathrm{out}}, \bm{q}_{n_\mathrm{in}})\\
        & \qquad\qquad \times\hat{I}_{\mi}^{(\ci)}(\bm{p}_\mathrm{filter} + \po, \bm{q}_{n_b}),\\
\end{split}
\end{align}
where the $p$-space filter coordinate $\bm{p}_\mathrm{filter} = -\Delta \bm{p} = \pin - \po$ is introduced, with values in the filter space
 $\mathcal{P}_\mathrm{filter} = \left\{ -P_\mathrm{filter}, \dots, 0, \dots, P_\mathrm{filter} \right\}^3 \subset \mathbb{Z}^3$,
and the $p$-space filter radius $P_\mathrm{filter} \in \mathbb{N}_0$ leading to a $p$-space filter size of $2 P_\mathrm{filter} + 1$.
The feature map $\hat{\bm{I}}$ is assumed to be zero for arguments outside its domain.
This corresponds to defining
an appropriate zero padding in a convolutional layer.
Note that for Eq.~\eqref{eq:def_discretized_layer_2} to hold, the radial basis filters need to be defined to be zero for $(\pin - \po) \notin \mathcal{P}_\mathrm{filter}$.

By rearranging the sums and introducing the kernel $\bm{K}$, containing all parameters and basis filters, which we propose to define as 
\begin{align}
\begin{split}
    \label{eq:discrete_kernel_def}
  K_{\mo, \mi}^{(\co, \ci)}(\bm{p}_\mathrm{filter}, \bm{q}_{n_\mathrm{out}}, \bm{q}_{n_\mathrm{in}}) \coloneqq &
    \sum_{\cf, k}W_{\ci, \cf, \co, k} \\
    & \qquad \times \sum_{\mf \in \{-\lf, \dots, \lf\}}
    C_{(\lf, \mf)(\li, \mi)}^{(\lo, \mo)} \\ & \qquad\qquad \times F_{\mf}^{(\cf, k)}(-\bm{p}_\mathrm{filter}, \bm{q}_{n_\mathrm{out}}, \bm{q}_{n_\mathrm{in}}),
\end{split}
\end{align}
Eq.~\eqref{eq:def_discretized_layer_2} can be further rewritten:
\begin{align}\label{eq:def_discretized_layer_3}
\begin{split}
  \hat{\mathcal{L}}_{\mo}^{(\co)}[\hat{\bm{I}}](\po, \bm{q}_{n_\mathrm{out}}) =&
    \sum_{\substack{\bm{q}_{n_\mathrm{in}} \in \mathcal{Q}_\mathrm{in}, \\ \ci, \\ \mi \in \{-\li, \dots, \li\}}}
    \sum_{\bm{p}_\mathrm{filter} \in \mathcal{P}_\mathrm{filter}}
  \sum_{\cf, k}W_{\ci, \cf, \co, k} \\
  & \qquad \times \sum_{\mf \in \{-\lf, \dots, \lf\}}
    C_{(\lf, \mf)(\li, \mi)}^{(\lo, \mo)} \\
    & \qquad\qquad \times F_{\mf}^{(\cf, k)}(-\bm{p}_\mathrm{filter}, \bm{q}_{n_\mathrm{out}}, \bm{q}_{n_\mathrm{in}})
        \hat{I}_{\mi}^{(\ci)}(\bm{p}_\mathrm{filter} + \po, \bm{q}_{n_\mathrm{in}}) \\
  =& \sum_{\substack{\bm{q}_{n_\mathrm{in}} \in \mathcal{Q}_\mathrm{in}, \\ \ci, \\ \mi \in \{-\li, \dots, \li\}}}
    \sum_{\bm{p}_\mathrm{filter} \in \mathcal{P}_\mathrm{filter}}
    K_{(\co, \mo), (\ci, \mi)}(\bm{p}_\mathrm{filter}, \bm{q}_{n_\mathrm{out}}, \bm{q}_{n_\mathrm{in}})
     \\ &\qquad \times \hat{I}_{\mi}^{(\ci)}(\bm{p}_\mathrm{filter} + \po, \bm{q}_{n_\mathrm{in}}).
\end{split}
\end{align}

The kernel $\bm{K}$ can be represented using an array of size
\begin{align}
    \dim\left(\mathcal{S}^{\bm{\tauo}}\right)\times
    \dim\left(\mathcal{S}^{\bm{\taui}}\right)\times
    P_\mathrm{filter} \times P_\mathrm{filter}  \times P_\mathrm{filter} \times
    Q_\mathrm{out} \times Q_\mathrm{in}.
\end{align}

\paragraph{Equivariance of Discretized Layer}
For discrete feature maps, property~\eqref{eq:dMRI_feature_map_action} needs to be adapted as follows:
\begin{align}\label{eq:dMRI_discrete_feature_map_action}
   (g, \mathcal{T}_{\bm{t}})[\hat{\bm{I}}](\bm{p}, \bm{q}) = \bm{D}^{\bm{\tau}}_g
   \Phi\left(\hat{\bm{I}}\left(\mathcal{\bm{R}}_{g^{-1}}(\bm{p}-\bm{t}),
                     \mathcal{\bm{R}}_{g^{-1}}\bm{q}\right)
              \right),
\end{align}
where $(g, \mathcal{T}_{\bm{t}}) \in \SE(3)$ with $g \in \SO(3)$ and $\bm{t} \in \mathbb{R}^3$,
$\bm{p} \in \mathcal{P}$, $\bm{q} \in \mathcal{Q}$, and $\Phi \colon (\mathbb{R}^3 \oplus \mathbb{R}^3 \to \mathcal{S}^{\bm{\tau}}) \to (\mathcal{P} \oplus \mathcal{Q} \to \mathcal{S}^{\bm{\tau}})$ is an interpolation.

Note that with Eq.~\eqref{eq:dMRI_discrete_feature_map_action} the equivariance property~\eqref{eq:conv_eq_property} is not
guaranteed for $\hat{\mathcal{L}}$ if the interpolation is not trivial, i.e.~if $\bm{\mathcal{R}}_{g^{-1}}(\bm{p}-\bm{t}) \not\in \mathcal{P}$
or $\bm{\mathcal{R}}_{g^{-1}}\bm{q} \not\in \mathcal{Q}$.
So equivariance may only hold approximately.

Also, because of discretization, aliasing artifacts may occur when using high angular filter orders $\lf$.
This means that high-order angular basis filters should not be used in filters. The maximum order depends on the filter size $P_\mathrm{f}$
(see~\citet{3dSteerableCNNs}).

\subsection{Efficient Implementation using 3D Convolutional Layer}\label{appendix:impl_3d_conv}
The layer defined in Eq.~\eqref{eq:def_discretized_layer_3} can be implemented using 3D convolutional layers (with computed kernels, see below), of which deep learning
frameworks like PyTorch~\citep{pytorch} support efficient implementations.
We will denote such a 3D convolutional layer as $\text{conv}_\text{3D}$.

In order to be used in a $\text{conv}_\text{3D}$ layer, the indices $c'_\mathrm{out}$ and $c'_\mathrm{in}$ are introduced
which allow indexing the kernel and the feature maps for all possible values of $(n_\mathrm{out}, \co, \mo)$ and $(n_\mathrm{in}, \ci, \mi)$, respectively,
using just single indices.
Also, the feature maps are reshaped to size
\begin{align}
    (\dim\left(\mathcal{S}^{\bm{\tau}}\right) \cdot Q) \times P_z \times P_y  \times P_x ,
\end{align}
and the kernel is reshaped to size
\begin{align}
    \left(Q_\mathrm{out} \cdot \dim\left(\mathcal{S}^{\bm{\tauo}}\right)\right) \times
    \left(Q_\mathrm{in} \cdot \dim\left(\mathcal{S}^{\bm{\taui}}\right)\right) \times
    P_\mathrm{filter} \times P_\mathrm{filter} \times P_\mathrm{filter}.
\end{align}

Using the reshaped feature map and kernel, the new indices $c'_\mathrm{out}, c'_\mathrm{in}$, and
replacing $\bm{q}_{n_\mathrm{out}}, \bm{q}_{n_\mathrm{in}}$ by their indices $n_\mathrm{out}, n_\mathrm{in}$, Eq.~\eqref{eq:def_discretized_layer_3} can be rewritten
to use the $\text{conv}_\text{3D}$ layer:
\begin{align}\label{eq:def_discretized_layer_conv3d}
\begin{split}
  \hat{\mathcal{L}}_{c'_\mathrm{out}(n_\mathrm{out}, \co, \mo)}[\hat{\bm{I}}](\po) ={}& \sum_{c'_\mathrm{in}} \sum_{\bm{p}_\mathrm{filter} \in \mathcal{P}_\mathrm{filter}}
     K_{c'_\mathrm{out}(n_\mathrm{out}, \co, \mo), c'_\mathrm{in}(n_\mathrm{in}, \ci, \mi)}(\bm{p}_\mathrm{filter}) \\
  & \qquad \times \hat{I}_{c'_\mathrm{in}(n_\mathrm{in}, \ci, \mi)}(\bm{p}_\mathrm{filter} + \po) \\
    ={}& \left(\sum_{c'_\mathrm{in}} \bm{K}_{c'_\mathrm{out}, c'_\mathrm{in}} \star
      \hat{\bm{I}}_{c'_\mathrm{in}}\right)(\po) \\
    ={}&\ \text{conv}_\text{3D}(\bm{K}, \hat{\bm{I}})_{c'_\mathrm{out}}(\po),
\end{split}
\end{align}
where $\star$ denotes cross-correlation.

\paragraph{Kernel Computation}
The computation of $\bm{K}$ can be split into multiple steps:
\begin{enumerate}
  \item Precomputation of Clebsch--Gordan coefficients $\bm{C}$
  \item Precomputation of angular basis $\bm{A}^{(\cf)}$ for all $\bm{p}_\mathrm{filter} \in \mathcal{P}_\mathrm{filter}$, $\bm{q}_{n_\mathrm{in}} \in \mathcal{Q}_\mathrm{in}$, and $\bm{q}_{n_\mathrm{out}} \in \mathcal{Q}_\mathrm{out}$ 
  \item (Pre)computation of radial basis $R^{(k)}$ for all $\bm{p}_\mathrm{filter} \in \mathcal{P}_\mathrm{filter}$, $\bm{q}_{n_\mathrm{in}} \in \mathcal{Q}_\mathrm{in}$, and $\bm{q}_{n_\mathrm{out}} \in \mathcal{Q}_\mathrm{out}$ 
  \item Multiplication of radial and angular basis as per Eq.~\eqref{eq:def_general_filter_kernel} and computation of $\bm{K}$ using learned weights $\bm{W}$ as per Eq.~\eqref{eq:discrete_kernel_def}
\end{enumerate}

As the angular basis does not contain any learned parameters, steps 1 and 2 can be computed before training.
Depending on the choice of radial basis function, the radial basis may or may not contain learned parameters.
If it does, only the inputs to the radial basis functions can be precomputed but the radial basis is computed
during training. If it does not contain learned parameters, then the whole radial basis can be precomputed.
Step 4 is always computed during training.

The number of radial basis functions used in the $p$-difference radial basis (Eq.~\eqref{eq:R_p_diff}) is chosen to be equal to $P_\mathrm{filter}$.
For the $q$-in (Eq.~\eqref{eq:R_q_in}) and $q$-out (Eq.~\eqref{eq:R_q_out}) radial bases, it is equal to the number of different lengths of $q$-vectors in $\mathcal{Q}_\mathrm{in}$ and $\mathcal{Q}_\mathrm{out}$, respectively.

Our implementation can be found at \url{https://github.com/philip-mueller/equivariant-deep-dmri}

\section{Vectors and Higher-Order Tensors as Inputs or Outputs}\label{appendix:higher_order_outputs}
As our proposed layer supports higher-order tensor inputs and outputs (e.g.\ vectors or order-2 tensors),
it can be used to build neural networks with such inputs or outputs,
e.g.\ networks for predicting diffusion tensors or using diffusion tensors as inputs.

For supporting vector ($l=1$) outputs, the output feature map of the output layer needs
to be configured to contain a single $l=1$ channel. No further changes
are required. Analogously the first layer can be configured to support vector inputs.
For supporting order-2 tensor ($l=2$) outputs an additional step is required after the output layer.
The layer uses spherical tensors in its feature maps, which for $l\le 1$ (scalars and vectors)
are (up to reordering) equal to Cartesian tensors but for $l>1$ are not.

In the case of $l>1$ a single Cartesian tensor can be decomposed into multiple spherical tensors.
For example, an order $l=2$ Cartesian tensors can be decomposed into one $l=0$, one $l=1$, and one $l=2$ spherical tensors,
corresponding to its isotropic, antisymmetric, and symmetric traceless parts, respectively~\citep{tensor_1,tensor_4}.
From this follows that in order to predict order-2 Cartesian tensors,
the output feature map of the output layer needs to 
contain one $l=0$, $l=1$, and $l=2$ channel each. These spherical tensors are then combined into
an order-2 Cartesian tensor (independently for each position).
If the predicted Cartesian tensor is known to be symmetric, then the $l=1$ spherical tensor is not required,
as it represents the antisymmetric part and is zero for symmetric tensors.
Analogously, for using an order-2 Cartesian tensor as input, the tensor is first decomposed into
one $l=0$, one $l=1$, and one $l=2$ spherical tensors, and these are then given to the first layer.

Note that it is also possible to predict (or use as input) multiple scalars, vectors, or order-2 tensors
or even a mix of different-order tensors (for each position).

\section{Proofs}\label{appendix:proofs}
\subsection{Proof of Equivariance Property~\texorpdfstring{\eqref{eq:conv_eq_property}}{(\ref{eq:conv_eq_property})} for Layer Definition~\texorpdfstring{\eqref{eq:conv_layer_definition}}{(\ref{eq:conv_layer_definition})}}
As each roto-translation can be built from a rotation followed
by a translation~\citep{rototrans_1}\label{appendix:layer_equi}
it is enough to prove equivariance under rotations and translations (of positions) independently.

\paragraph{Proof of Equivariance to Joint Rotations in $p$- and $q$-space}
To be proven is the following property on Eq.~\eqref{eq:conv_layer_definition}:
\begin{align}\label{eq:rot_prop_of_layer}
  (\mathcal{L} \circ g)[\bm{I}](\bm{p}, \bm{q}) = (g \circ \mathcal{L})[\bm{I}](\bm{p}, \bm{q})
\end{align}
for $g \in \SO(3)$, where $g$ is applied as:
\begin{align}\label{eq:apply_rot_to_layer}
  g[\bm{I}](\bm{p}, \bm{q}) := \bm{D}^{\bm{\tau}}_g\bm{I}\left(\bm{\mathcal{R}}_{g^{-1}}\bm{p}, \bm{\mathcal{R}}_{g^{-1}}\bm{q}\right).
\end{align}

Eq.~\eqref{eq:rot_prop_of_layer} is proven for each output channel $\co$ and output spherical tensor component $\mo$ independently.
For the proof, first definition~\eqref{eq:conv_layer_definition} is inserted into the left-hand side of Eq.~\eqref{eq:rot_prop_of_layer} (for a single $\co, \mo$) and then $g$ is applied using Eq.~\eqref{eq:apply_rot_to_layer}.
In the next step, we substitute $\po \mapsto \bm{\mathcal{R}}_g\po'$, $\pin \mapsto \bm{\mathcal{R}}_g\pin'$,
$\qo \mapsto \bm{\mathcal{R}}_g\qo'$, and $\qi \mapsto \bm{\mathcal{R}}_g\qi'$ and then rewrite to sum over $\pin'$ and $\qi'$.
After that, the equivariance property~\eqref{eq:kernel_eq_property} of $\bm{F}$ is used.
Then the equivariance property~\eqref{eq:TP_equivariance} of the tensor product is applied.
Finally, Eq.~\eqref{eq:conv_layer_definition} is used again and $\po, \qo$ are back-substituted resulting in the right-hand side of Eq.~\eqref{eq:rot_prop_of_layer} (for a single $\co, \mo$).
In detail:
{
\allowdisplaybreaks
\begin{align*}
  &(\mathcal{L} \circ g)[\bm{I}]^{(\co)}_{\mo}(\po, \qo)\\
& \qquad = \sum_{\ci, \cf, k}W^{(\cf)}_{\ci, \co, k}
     \sum_{\substack{\mf \in \{-\lf, \dots, \lf \}, \\ \mi \in \{-\li, \dots, \li \}}}
     C_{(\lf, \mf)(\li, \mi)}^{(\lo, \mo)}\\
  &\qquad\qquad \times \sum_{\substack{\pin \in \mathbb{R}^3, \\ \qi \in \mathbb{R}^3}}
  F_{\mf}^{(\cf, k)}(\po - \pin, \qo, \qi)
    g[\bm{I}]_{\mi}^{(\ci)}(\pin, \qi) \\ % definition
& \qquad= \sum_{\ci, \cf, k}W^{(\cf)}_{\ci, \co, k}
     \sum_{\substack{\mf \in \{-\lf, \dots, \lf \}, \\ \mi \in \{-\li, \dots, \li \}}}
     C_{(\lf, \mf)(\li, \mi)}^{(\lo, \mo)}\\
  &\qquad\qquad \times \sum_{\substack{\pin \in \mathbb{R}^3, \\ \qi \in \mathbb{R}^3}}
  F_{\mf}^{(\cf, k)}(\po - \pin, \qo, \qi) \\
    & \qquad\qquad \times \sum_{\mip \in \{-\li, \dots, \li\}} \left(D^{(\li)}_g\right)_{\mi, \mip}
    \bm{I}_{\mip}^{(\ci)}\left(\bm{\mathcal{R}}_{g^{-1}}\pin, \bm{\mathcal{R}}_{g^{-1}}\qi\right) \\% apply g
& \qquad= \sum_{\ci, \cf, k}W^{(\cf)}_{\ci, \co, k}
     \sum_{\substack{\mf \in \{-\lf, \dots, \lf \}, \\ \mi \in \{-\li, \dots, \li \}}}
     C_{(\lf, \mf)(\li, \mi)}^{(\lo, \mo)} \\
      & \qquad\qquad \times \sum_{\substack{\bm{\mathcal{R}}_g\pin' \in \mathbb{R}^3, \\ \bm{\mathcal{R}}_g\qi' \in \mathbb{R}^3}}
      F_{\mf}^{(\cf, k)}(\bm{\mathcal{R}}_g\po' - \bm{\mathcal{R}}_g\pin', \bm{\mathcal{R}}_g\qo', \bm{\mathcal{R}}_g\qi') \\
      & \qquad\qquad \times\sum_{\mip \in \{-\li, \dots, \li\}} \left(D^{(\li)}_g\right)_{\mi, \mip}
      \bm{I}_{\mip}^{(\ci)}(\pin', \qi') \\ % a -> R a'
& \qquad= \sum_{\ci, \cf, k}W^{(\cf)}_{\ci, \co, k}
     \sum_{\substack{\mf \in \{-\lf, \dots, \lf \}, \\ \mi \in \{-\li, \dots, \li \}}}
     C_{(\lf, \mf)(\li, \mi)}^{(\lo, \mo)} \\
          & \qquad\qquad \times\sum_{\substack{\pin' \in \mathbb{R}^3, \\ \qi' \in \mathbb{R}^3}}
          F_{\mf}^{(\cf, k)}(\bm{\mathcal{R}}_g(\po' - \pin'), \bm{\mathcal{R}}_g\qo', \bm{\mathcal{R}}_g\qi') \\
          & \qquad\qquad \times\sum_{\mip \in \{-\li, \dots, \li\}} \left(D^{(\li)}_g\right)_{\mi, \mip}
      \bm{I}_{\mip}^{(\ci)}(\pin', \qi') \\ % reformulate sum
& \qquad= \sum_{\ci, \cf, k}W^{(\cf)}_{\ci, \co, k}
        \sum_{\substack{\mf \in \{-\lf, \dots, \lf \}, \\ \mi \in \{-\li, \dots, \li \}}}
          C_{(\lf, \mf)(\li, \mi)}^{(\lo, \mo)} \\
          & \qquad\qquad \times\sum_{\substack{\pin' \in \mathbb{R}^3, \\ \qi' \in \mathbb{R}^3}}
          \Bigg(\sum_{\mfp \in \{-\lf, \dots, \lf\}} \left(D^{(\lf)}_g\right)_{\mf, \mfp} \\ & \qquad\qquad\qquad\qquad\qquad \times F_{\mfp}^{(\cf, k)}(\po' - \pin', \qo', \qi')\Bigg) \\
          & \qquad\qquad \times\Bigg(\sum_{\mip \in \{-\li, \dots, \li \}} \left(D^{(\li)}_g\right)_{\mi, \mip}
          \bm{I}_{\mip}^{(\ci)}(\pin', \qi')\Bigg) \\ % equivariance of F
  & \qquad= \sum_{\ci, \cf, k}W^{(\cf)}_{\ci, \co, k}
        \sum_{\mop \in \{-\lo, \dots, \lo\}}\left(D^{(\lo)}_g\right)_{\mo, \mop}\\
        &\qquad\qquad \times \sum_{\substack{\mf \in \{-\lf, \dots, \lf\}, \\ \mi \in \{-\li, \dots, \li\}}}
        C_{(\lf, \mf)(\li, \mi)}^{(\lo, \mop)} \\
        & \qquad\qquad \times\sum_{\substack{\pin' \in \mathbb{R}^3, \\ \qi' \in \mathbb{R}^3}}
        F_{\mf}^{(\cf, k)}(\po' - \pin', \qo', \qi')
        \bm{I}_{\mi}^{(\ci)}(\pin', \qi') \\ % equivariance of TP
      & \qquad= \sum_{\mop \in \{-\lo, \dots, \lo\}}\left(D^{(\lo)}_g\right)_{\mo, \mop}
       \mathcal{L}[\bm{I}]^{(\co)}_{\mop}(\po', \qo') \\
      & \qquad= \sum_{\mop \in \{-\lo, \dots, \lo\}}\left(D^{(\lo)}_g\right)_{\mo, \mop}\\
       &\qquad\qquad \times \mathcal{L}[\bm{I}]^{(\co)}_{\mop}\left(\bm{\mathcal{R}}_{g^{-1}}\po, \bm{\mathcal{R}}_{g^{-1}}\qo\right) \\
      & \qquad= (g \circ \mathcal{L})[\bm{I}]^{(\co)}_{\mo}(\po, \qo).
\end{align*}

\paragraph{Proof of Equivariance to Translations in $p$-space}
To be proven is the following property on Eq.~\eqref{eq:conv_layer_definition}:
\begin{align}\label{eq:trans_prop_of_layer}
  (\mathcal{L} \circ \mathcal{T}_{\bm{t}})[\bm{I}](\bm{p}, \bm{q}) = (\mathcal{T}_{\bm{t}} \circ \mathcal{L})[\bm{I}](\bm{p}, \bm{q})
\end{align}
for $\bm{t} \in \mathbb{R}^3$, where $\mathcal{T}_{\bm{t}}$ is applied as:
\begin{align}\label{eq:apply_trans_to_layer}
  \mathcal{T}_{\bm{t}}[\bm{I}](\bm{p}, \bm{q}) := \bm{I}(\bm{p} - \bm{t}, \bm{q}).
\end{align}

Eq.~\eqref{eq:trans_prop_of_layer} is proven for each output channel $\co$ and output spherical tensor component $\mo$ independently.
For the proof, first the definition~\eqref{eq:conv_layer_definition} is inserted
into the left-hand side of Eq.~\eqref{eq:trans_prop_of_layer} (for a single $\co, \mo$) and then $\mathcal{T}_{\bm{t}}$ is applied using Eq.~\eqref{eq:apply_trans_to_layer}.
In the next step, we substitute $\po \mapsto \po' + \bm{t}$ and $\pin \mapsto \pin' + \bm{t}$.
Then Eq.~\eqref{eq:conv_layer_definition} is used again and $\po$ is back-substituted resulting in the right-hand side of Eq.~\eqref{eq:trans_prop_of_layer} (for a single $\co, \mo$). In detail:
\allowdisplaybreaks
\begin{align*}
  &(\mathcal{L} \circ \mathcal{T}_{\bm{t}})[\bm{I}]^{(\co)}_{\mo}(\po, \qo) \\
& \qquad = \sum_{\ci, \cf, k}W^{(\cf)}_{\ci, \co, k}
     \sum_{\substack{\mf \in \{-\lf, \dots, \lf \}, \\ \mi \in \{-\li, \dots, \li \}}}
     C_{(\lf, \mf)(\li, \mi)}^{(\lo, \mo)}\\
    &\qquad\qquad \times\sum_{\substack{\pin \in \mathbb{R}^3, \\ \qi \in \mathbb{R}^3}}
    F_{\mf}^{(\cf, k)}(\po - \pin, \qo, \qi)
     \mathcal{T}_{\bm{t}}[\bm{I}]_{\mi}^{(\ci)}(\pin, \qi) \\ % definition
& \qquad =  \sum_{\ci, \cf, k}W^{(\cf)}_{\ci, \co, k}
     \sum_{\substack{\mf \in \{-\lf, \dots, \lf \}, \\ \mi \in \{-\li, \dots, \li \}}}
     C_{(\lf, \mf)(\li, \mi)}^{(\lo, \mo)}\\
     &\qquad\qquad \times \sum_{\substack{\pin \in \mathbb{R}^3, \\ \qi \in \mathbb{R}^3}}
     F_{\mf}^{(\cf, k)}(\po - \pin, \qo, \qi)
      I_{\mi}^{(\ci)}(\pin - \bm{t}, \qi) \\ % apply T
& \qquad =  \sum_{\ci, \cf, k}W^{(\cf)}_{\ci, \co, k}
     \sum_{\substack{\mf \in \{-\lf, \dots, \lf \}, \\ \mi \in \{-\li, \dots, \li \}}}
     C_{(\lf, \mf)(\li, \mi)}^{(\lo, \mo)}\\
     & \qquad\qquad \times\sum_{\substack{\pin' \in \mathbb{R}^3, \\ \qi \in \mathbb{R}^3}}
     F_{\mf}^{(\cf, k)}(\po' - \pin', \qo, \qi)
     I_{\mi}^{(\ci)}(\pin', \qi) \\% p -> p' + t
& \qquad =  \mathcal{L}[\bm{I}]^{(\co)}_{\mo}(\po', \qo) \\ % definition
& \qquad =  \mathcal{L}[\bm{I}]^{(\co)}_{\mo}(\po - \bm{t}, \qo) \\ % back substitute
& \qquad =  (\mathcal{T}_{\bm{t}} \circ \mathcal{L})[\bm{I}]^{(\co)}_{\mo}(\po, \qo).
\end{align*}
}

\subsection{Proof of Equivariance Property~\eqref{eq:kernel_eq_property} for Filter Basis defined in Eq.~\texorpdfstring{\eqref{eq:def_general_filter_kernel}}{(\ref{eq:def_general_filter_kernel})}}\label{appendix:filter_basis_equi}
For the proof, first definition~\eqref{eq:def_general_filter_kernel} is inserted into the left-hand side of Eq.~\eqref{eq:kernel_eq_property}.
After that, the invariance property~\eqref{eq:R_invariance} of $R^{(k)}$ and the equivariance property~\eqref{eq:A_equivariance}
of $\bm{A}^{(\cf)}$ are used.
Then we make use of the commutativity of scalar multiplication (as $R^{(k)}$ is scalar-valued) and
finally the definition~\eqref{eq:def_general_filter_kernel} is used again resulting in the right-hand side of Eq.~\eqref{eq:kernel_eq_property}.
In detail:
\begin{align*}
  \bm{F}^{(\cf, k)}(\bm{\mathcal{R}}_g\Delta \bm{p}, \bm{\mathcal{R}}_g\qo, \bm{\mathcal{R}}_g\qi)
  ={} &R^{(k)}(\bm{\mathcal{R}}_g\Delta \bm{p}, \bm{\mathcal{R}}_g\qo, \bm{\mathcal{R}}_g\qi)\\
    &\qquad \times \bm{A}^{(\cf)}(\bm{\mathcal{R}}_g\Delta \bm{p}, \bm{\mathcal{R}}_g\qo, \bm{\mathcal{R}}_g\qi)\\
 ={} &R^{(k)}(\Delta \bm{p}, \qo, \qi)\\
 &\qquad \times \bm{A}^{(\cf)}(\bm{\mathcal{R}}_g\Delta \bm{p}, \bm{\mathcal{R}}_g\qo, \bm{\mathcal{R}}_g\qi) \\
={} &R^{(k)}(\Delta \bm{p}, \qo, \qi) \bm{D}^{(\lf)}_g \bm{A}^{(\cf)}(\Delta \bm{p}, \qo, \qi) \\
={} &\bm{D}^{(\lf)}_g R^{(k)}(\Delta \bm{p}, \qo, \qi) \bm{A}^{(\cf)}(\Delta \bm{p}, \qo, \qi) \\
={} &\bm{D}^{(\lf)}_g\bm{F}^{(\cf, k)}(\Delta \bm{p}, \qo, \qi)
\end{align*}

\subsection{Proof of Invariance Property~\eqref{eq:R_invariance} for Radial Bases defined in Eq.~\eqref{eq:R_p_diff}, Eq.~\eqref{eq:R_q_in}, Eq.~\eqref{eq:R_q_out}, and Eq.~\eqref{eq:R_prod}}\label{appendix:proof_radial_basis_inv}
The invariances, as defined in Eq.~\eqref{eq:R_invariance}, for the $p$-difference (Eq.~\eqref{eq:R_p_diff}), $q$-in (Eq.~\eqref{eq:R_q_in}), and $q$-out (Eq.~\eqref{eq:R_q_out}) radial bases follow from the invariance of the $L_2$ norm.

So it is left to prove Eq.~\eqref{eq:R_invariance} for the product of radial bases as defined in Eq.~\eqref{eq:R_prod}:
\begin{align*}
 R_{\text{prod}}^{(k_1, k_2)}(\bm{\mathcal{R}}_g\Delta \bm{p}, \bm{\mathcal{R}}_g\qo, \bm{\mathcal{R}}_g\qi)
 ={} &R^{(k_1)}_1(\bm{\mathcal{R}}_g\Delta \bm{p}, \bm{\mathcal{R}}_g\qo, \bm{\mathcal{R}}_g\qi)\\
  &\qquad \times R^{(k_2)}_2(\bm{\mathcal{R}}_g\Delta \bm{p}, \bm{\mathcal{R}}_g\qo, \bm{\mathcal{R}}_g\qi) \\
={} &R^{(k_1)}_1(\Delta \bm{p}, \qo, \qi) R^{(k_2)}_2(\Delta \bm{p}, \qo, \qi) \\
 ={} &R_{\text{prod}}^{(k_1, k_2)}(\Delta \bm{p}, \qo, \qi),
\end{align*}
where the invariance property~\eqref{eq:R_invariance} of the radial bases $R^{(k_1)}_1$ and $R^{(k_2)}_2$ being combined is used.

\subsection{Proof of Equivariance Property~\eqref{eq:A_equivariance} for Angular Bases defined in Eq.~\eqref{eq:angular_basis_p}, Eq.~\eqref{eq:angular_basis_q}, Eq.~\eqref{eq:angular_basis_pq}, and Eq.~\eqref{eq:angular_combined_def}}\label{appendix:proof_angular_basis_equi}
For the proof of property~\eqref{eq:A_equivariance} for Eq.~\eqref{eq:angular_basis_p}, first the definition of $\bm{A}^{(\cf)}_{p\text{-diff}}$~\eqref{eq:angular_basis_p} is inserted for $\bm{A}^{(\cf)}$ into the left-hand side of Eq.~\eqref{eq:A_equivariance}, then the invariance of the $L_2$ norm is used.
Finally, the equivariance~\eqref{eq:real_spherical_harmonics_rotation_1} of the spherical harmonics
is applied and definition~\eqref{eq:angular_basis_p} is used again for $\bm{A}^{(\cf)}$ resulting in the right-hand side of Eq.~\eqref{eq:A_equivariance}. In detail:
\begin{align*}
\bm{A}^{(\cf)}_{p\text{-diff}}(\bm{\mathcal{R}}_g\Delta \bm{p}, \bm{\mathcal{R}}_g\qo, \bm{\mathcal{R}}_g\qi)
    ={} &\bm{Y}^{(\lf)}\left(
\frac{\bm{\mathcal{R}}_g\Delta \bm{p}}
{\left\lVert\bm{\mathcal{R}}_g\Delta \bm{p}\right\rVert_2}\right) \\
={} &\bm{Y}^{(\lf)}\left(\bm{\mathcal{R}}_g
\frac{\Delta \bm{p}}
{\left\lVert\Delta \bm{p}\right\rVert_2}\right) \\
={} &\bm{D}^{(\lf)}_g\bm{Y}^{(\lf)}\left(
\frac{\Delta \bm{p}}
{\left\lVert\Delta \bm{p}\right\rVert_2}\right) \\
={} &\bm{D}^{(\lf)}_g\bm{A}_{p\text{-diff}}^{(\cf)}(\Delta \bm{p}, \qo, \qi),
\end{align*}

The proofs for Eq.~\eqref{eq:angular_basis_q} and Eq.~\eqref{eq:angular_basis_pq} are analog but additionally the distributive law of matrix multiplication and vector addition is used, i.e.
\begin{align*}
    \bm{\mathcal{R}}_g\qo - \bm{\mathcal{R}}_g\qi = \bm{\mathcal{R}}_g(\qo - \qi),
\end{align*}
and
\begin{align*}
    \bm{\mathcal{R}}_g\Delta \bm{p} - (\bm{\mathcal{R}}_g\qo - \bm{\mathcal{R}}_g\qi)
    ={} &\bm{\mathcal{R}}_g\Delta \bm{p} - \bm{\mathcal{R}}_g(\qo - \qi) \\
    ={} &\bm{\mathcal{R}}_g(\Delta \bm{p} - (\qo - \qi)),
\end{align*}
respectively.

For the proof of property~\eqref{eq:A_equivariance} for the combined angular basis defined in Eq.~\eqref{eq:angular_combined_def}, first the definition of $\bm{A}^{(\cf)}_\text{TP}$~\eqref{eq:angular_combined_def} is inserted for $\bm{A}^{(\cf)}$ into the left-hand side of Eq.~\eqref{eq:A_equivariance} for each angular filter channel $\cf$ and spherical tensor component $\mf$ independently.
Then the equivariance property~\eqref{eq:A_equivariance} of the angular bases $\left(\bm{A_1}\right)$ and
$\left(\bm{A_2}\right)$ is used.
Finally, the equivariance~\eqref{eq:TP_equivariance} of the tensor product is applied and definition~\eqref{eq:angular_combined_def}  is used again for $\bm{A}^{(\cf)}$ resulting in the right-hand side of Eq.~\eqref{eq:A_equivariance}.
In detail:
\begin{align*}
  &\left(A_\text{TP}\right)^{(\cf)}_{\mf}(\bm{\mathcal{R}}_g\Delta \bm{p}, \bm{\mathcal{R}}_g\qo, \bm{\mathcal{R}}_g\qi)\\
&\qquad =  \sum_{\substack{m_1 \in \{-l_1, \dots, l_1\}, \\ m_2 \in \{-l_2, \dots, l_2\}}}
  C_{(l_1, m_1)(l_2, m_2)}^{(\lf, \mf)} \\
  &\qquad\qquad \times \left(A_1\right)^{(c_1)}_{m_1}(\bm{\mathcal{R}}_g\Delta \bm{p}, \bm{\mathcal{R}}_g\qo, \bm{\mathcal{R}}_g\qi)
   \left(A_2\right)^{(c_2)}_{m_2}(\bm{\mathcal{R}}_g\Delta \bm{p}, \bm{\mathcal{R}}_g\qo, \bm{\mathcal{R}}_g\qi) \\
&\qquad = \sum_{\substack{m_1 \in \{-l_1, \dots, l_1\}, \\ m_2 \in \{-l_2, \dots, l_2\}}}
  C_{(l_1, m_1)(l_2, m_2)}^{(\lf, \mf)} \\
  &\qquad\qquad \times  \Bigg[\sum_{m'_1 \in \{-l_1, \dots, l_1\}}
    D^{(l_1)}_{m_1, m'_1} \left(A_1\right)_{m'_1}^{(c_1)}(\Delta \bm{p}, \qo, \qi) \Bigg] \\
  &\qquad\qquad \times  \Bigg[\sum_{m'_2 \in \{-l_2, \dots, l_2\}}
    D^{(l_2)}_{m_2, m'_2}  \left(A_2\right)_{m'_2}^{(c_2)}(\Delta \bm{p}, \qo, \qi) \Bigg] \\
&\qquad = \sum_{\mfp \in \{-\lf, \dots, \lf\}}D^{(\lf)}_{\mf, \mfp}
\sum_{\substack{m_1 \in \{-l_1, \dots, l_1\}, \\ m_2 \in \{-l_2, \dots, l_2\}}}
C_{(l_1, m_1)(l_2, m_2)}^{(\lf, \mfp)} \\
  &\qquad\qquad \times  \left(A_1\right)^{(c_1)}_{m_1}(\Delta \bm{p}, \qo, \qi)
  \left(A_2\right)^{(c_2)}_{m_2}(\Delta \bm{p}, \qo, \qi) \\
&\qquad = \sum_{\mfp \in \{-\lf, \dots, \lf\}}D^{(\lf)}_{\mf, \mfp} \left(A_\text{TP}\right)^{(\cf)}_{\mf}(\Delta \bm{p}, \qo, \qi).
\end{align*}

\section{Layer and Channel Configurations}
\label{appendix:architectures}
\subsection{Equivariant Models}
\label{appendix:equi_architectures}

\begin{table}[ht]
    \centering
    \renewcommand{\arraystretch}{1.2}
    \caption{Channel configurations of our equivariant models with \texttt{late} $q$-reduction. The tuples $(a,b,c,d)$ are the $\tauo$ of each layer, i.e.\ they describe the output channels of a layer, where $a, b, c, d$ are the numbers of channels of outputs orders ($\lo$) 0, 1, 2, and 3, respectively, e.g.\ $(4, 3, 2, 1)$ means that a layers has four order 0 (scalar), three order 1 (vector), two order 2, and one order 3 output channels.
    The column ``in'' describes the input feature map of the first layer, while the subsequent columns describe the output feature maps of the layers: one $pq$-layer, one $q$-reduction layer, and up to five $p$-layers. The ``Layers'' column describes the number of $pq$-layers, $q$-reduction layers (always 1), and $p$-layers, where the notation $(l2)$ is used for layers with additional $\lo=2$ channels, and $(l2/l3)$ for layers with additional $\lo=2$ and $\lo=3$ channels. The row ``$Q$'' describes the number of sampled $q$-space coordinates in the respective feature maps. Note that there is no $q$-space in the output of the $q$-reduction layer and the $p$-layers. 
    The shown channel configurations are the results of hyperparameter tuning for each given layer configuration (some confgurations with many channels had to be excluded from hyperparameter tuning, as they did not fit into GPU memory). Note that the channel configurations do not differ for different filter bases or $p$ radial bases.}\label{tab:same_q}
    \scriptsize
    \tabcolsep=0.1cm
    \begin{tabular}{@{}lllllllll@{}}
    \toprule
    Layers & in & $pq$-1 & $q$-reduction & $p$-1 & $p$-2 & $p$-3 & $p$-4 & $p$-5 \\ \midrule
    1+1+2  & $(1,0,0,0)$ & $(5,3,0,0)$ & $(5,3,0,0)$ & $(10,5,0,0)$ & $(1,0,0,0)$ & -         & -         & - \\
    1+1+3  & $(1,0,0,0)$& $(5,3,0,0)$ & $(5,3,0,0)$ & $(50,10,0,0)$ & $(20,5,0,0)$ & $(1,0,0,0)$ & -         & - \\
    1+1+4  & $(1,0,0,0)$& $(7,4,0,0)$ & $(7,4,0,0)$ & $(20,5,0,0)$ & $(10,3,0,0)$ & $(5,2,0,0)$ & $(1,0,0,0)$ & - \\
    1+1+4($l2$)  & $(1,0,0,0)$& $(7,4,0,0)$ & $(7,4,0,0)$ & $(20,5,2,0)$ & $(10,3,1,0)$ & $(5,2,0,0)$ & $(1,0,0,0)$ & - \\
    1+1+4($l2/l3$)  & $(1,0,0,0)$& $(7,4,0,0)$ & $(7,4,0,0)$ & $(20,5,2,1)$ & $(10,3,1,0)$ & $(5,2,0,0)$ & $(1,0,0,0)$ & - \\
    1($l2$)+1+4($l2$)  & $(1,0,0,0)$& $(5,3,1,0)$ & $(5,3,1,0)$ & $(20,5,2,0)$ & $(10,3,1,0)$ & $(5,2,0,0)$ & $(1,0,0,0)$ & - \\
    1($l2/l3$)+1+4($l2/l3$)  & $(1,0,0,0)$& $(5,3,1,1)$ & $(5,3,1,1)$ & $(20,5,2,1)$ & $(10,3,1,0)$ & $(5,2,0,0)$ & $(1,0,0,0)$ & - \\
    1+1+5 & $(1,0,0,0)$ & $(5,3,0,0)$ & $(5,3,0,0)$ & $(20,5,0,0)$ & $(10,3,0,0)$ & $(5,2,0,0)$ & $(3,1,0,0)$ & $(1,0,0,0)$ \\
    \midrule 
    $Q$ & 42 & 42 & - & - & - & - & - \\
    \bottomrule
    \end{tabular}
\end{table}

\begin{table}[ht]
    \centering
    \renewcommand{\arraystretch}{1.2}
    \caption{Like Table~\ref{tab:same_q}, but for \texttt{gradual} $q$-reduction.
    }\label{tab:decr_q}
    \scriptsize
    \tabcolsep=0.1cm
    \begin{tabular}{@{}llllllll@{}}
    \toprule
    Layers & in & $pq$-1 & $pq$-2 & $q$-reduction & $p$-1 & $p$-2 & $p$-3 \\ \midrule
    0+1+3 & $(1,0,0,0)$ & - & - & $(100,20,0,0)$ & $(50,10,0,0)$ & $(10,5,0,0)$ & $(1,0,0,0)$ \\
    1+1+2  & $(1,0,0,0)$ & -         & $(15,7,0,0)$ & $(70,10,0,0)$ & $(20,5,0,0)$ & $(1,0,0,0)$ & - \\
    1+1+3  & $(1,0,0,0)$ & -         & $(15,7,0,0)$ & $(70,10,0,0)$ & $(20,5,0,0)$ & $(10,3,0,0)$ & $(1,0,0,0)$ \\
    2+1+1  & $(1,0,0,0)$ & $(3,2,0,0)$ & $(5,3,0,0)$ & $(70,10,0,0)$ & $(1,0,0,0)$ & -         & - \\
    \midrule 
    $Q$ & 42 & 27 & 7 & - & - & - & - \\
    \bottomrule
    \end{tabular}
\end{table}

\clearpage
\subsection{Non-Rotation-Equivariant Models}
\label{appendix:reference_architectures}
\begin{table}[ht]
    \centering
    \renewcommand{\arraystretch}{1.2}
    \caption{Channel configurations of non-rotation-equivariant models used as reference.
    We experimented with different numbers of channels but found 
that small changes in the number of channels did not affect the quality much. The shown channel configurations are the best we found for each number of layers and we assume that they are very near the optimum for the given training setup.
Also note that using similar feature map sizes as used in the equivariant models (and thus having much more channels and parameters) leads to much worse results.}\label{tab:non_equi}
    \scriptsize
    \tabcolsep=0.1cm
    \begin{tabular}{@{}llllllll@{}}
    \toprule
    Layers & in & layer-1 & layer-2 & layer-3 & layer-4 & layer-5 & layer-6 \\ \midrule
    3 {\tiny(few channels)}  & 46 & 5 & 3 & 1 & - & - & -\\
    3 {\tiny(many channels)}  & 46 & 10 & 5 & 1 & - & - & -\\ 
    4 {\tiny(few channels)} & 46 & 15 & 5 & 3 & 1 & - & -\\
    4 {\tiny(many channels)} & 46 & 30 & 10 & 5 & 1 & - & -\\
    5 {\tiny(few channels)} & 46 & 5 & 15 & 5 & 3 & 1 & -\\
    5 {\tiny(many channels)} & 46 & 10 & 30 & 10 & 5 & 1 & - \\
    6 {\tiny(few channels)} & 46 & 5 & 5 & 15 & 5 & 3 & 1 \\ 
    6 {\tiny(many channels)} & 46 & 5 & 10 & 30 & 10 & 5 & 1 \\ 
    6 {\tiny(matched feature maps (fm) small)} & 46 & 378 & 119 & 95 & 160 & 80 & 1 \\
    6 {\tiny(matched feature maps (fm) large)} & 46 & 378 & 175 & 180 & 160 & 80 & 1 \\ 
    \bottomrule
    \end{tabular}
\end{table}

\end{document}